\documentclass[runningheads]{llncs}
\usepackage[T1]{fontenc}
\usepackage{graphicx}
\usepackage[colorlinks=true]{hyperref}

\usepackage{amsfonts}       
\usepackage{nicefrac}       
\usepackage{microtype}      
\usepackage{lipsum}
\usepackage{rotating}
\usepackage{float}
\usepackage{multirow,epsfig,pdflscape,graphicx,amsmath,amssymb}
\usepackage{xspace}
\def\ood{\texttt{OOD}\xspace}
\def\iid{\texttt{IID}\xspace}
\def\id{\texttt{ID}\xspace}
\def\sota{\texttt{SOTA}\xspace}
\def\dnn{\texttt{DNN}\xspace}

\def\X{{\mathcal{X}}}
\def\Y{{\mathcal{Y}}}
\def\D{{\mathcal{D}}}

\def\cP{{\mathcal{P}}}
\def\cQ{{\mathcal{Q}}}
\def\cM{{\mathcal{M}}}

\def\lce{{\mathcal{L}_{\rm CE}}}

\DeclareMathOperator*{\argmax}{arg\,max}
\DeclareMathOperator*{\minimize}{minimize}
\def\etal{{et al.\xspace}}
\def\xt{{\Tilde{x}}}

\newcommand{\vertext}[1]{\begin{turn}{90} {#1} \end{turn}}
\usepackage{array, tabularx,  ragged2e,  booktabs}

\usepackage{amsfonts}       

\usepackage{comment}
\usepackage{caption}
\usepackage{subcaption}

\usepackage{nicefrac}       
\usepackage{microtype}      
\usepackage{lipsum}
\usepackage{rotating}
\usepackage{float}
\usepackage{multirow,epsfig,pdflscape,graphicx,amsmath,amssymb}
\usepackage{xspace}
\def\ood{\texttt{OOD}\xspace}
\def\iid{\texttt{IID}\xspace}
\def\id{\texttt{ID}\xspace}
\def\sota{\texttt{SOTA}\xspace}
\def\dnn{\texttt{DNN}\xspace}

\def\X{{\mathcal{X}}}
\def\Y{{\mathcal{Y}}}
\def\D{{\mathcal{D}}}

\def\cP{{\mathcal{P}}}
\def\cQ{{\mathcal{Q}}}
\def\cM{{\mathcal{M}}}

\def\lce{{\mathcal{L}_{\rm CE}}}
\def\etal{{et al.\xspace}}
\def\xt{{\Tilde{x}}}

\usepackage{array, tabularx,  ragged2e,  booktabs}
\usepackage[colorlinks=true]{hyperref}
\usepackage[linesnumbered,ruled,vlined]{algorithm2e}

\begin{document}
\title{A Novel Data Augmentation Technique for Out-of-Distribution Sample Detection using Compounded Corruptions}
%
%
\titlerunning{OOD Sample Detection using Compounded Corruptions}

\author{
	Ramya Hebbalaguppe$^{1,2}$ 
	\quad Soumya Suvra Ghosal$^{1}$ 
	\quad Jatin Prakash$^{1}$ \\
	\quad Harshad Khadilkar$^{2}$
	\quad Chetan Arora$^1$ 
	\\ 
	$^1$IIT Delhi, India 
	\quad$^2$TCS Research, India\thanks{This is the extended version of ECML'22 submission that includes detailed description and experiments}
	\\
	{\small \url{https://github.com/cnc-ood}}
}
\authorrunning{R. Hebbalaguppe et al.}
%
\institute{}
%
\maketitle              
\begin{abstract}
Modern deep neural network models are known to erroneously classify out-of-distribution (\ood) test data into one of the in-distribution (\id) training classes with high confidence. This can have disastrous consequences for safety-critical applications. A popular mitigation strategy is to train a separate classifier that can detect such \ood samples at test time. In most practical settings \ood examples are not known at train time, and hence a key question is: \emph{how to augment the \id data with synthetic \ood samples for training such an \ood detector?} In this paper, we propose a novel \textbf{C}ompou\textbf{n}ded \textbf{C}orruption (CnC) technique for the \ood data augmentation. One of the major advantages of CnC is that it does not require any hold-out data apart from training set. Further, unlike current state-of-the-art (\sota) techniques, CnC does not require backpropagation or ensembling at the test time, making our method much faster at inference. Our extensive comparison with $20$ methods from the major conferences in last 4 years show that a model trained using CnC based data augmentation, significantly outperforms \sota, both in terms of \ood detection accuracy as well as inference time. We include a detailed post-hoc analysis to investigate the reasons for the success of our method and identify higher relative entropy and diversity of CnC samples as probable causes. Theoretical insights via a piece-wise decomposition analysis on a two-dimensional dataset to reveal (visually and quantitatively) that our approach leads to a tighter boundary around ID classes, leading to better detection of \ood samples. 
	
\end{abstract}

\section{Introduction}
\label{sec:intro}

Deep neural network (\dnn) models generalize well when the test data is independent and identically distributed (\iid) with respect to training data~\cite{simonyan2014very}. However, the condition is difficult to enforce in the real world due to distributional drifts, covariate shift, and/or adversarial perturbations. A \emph{reliable} system based on a \dnn model must be able to detect an \ood sample, and either abstain from making any decision on such samples, or flag them for human intervention. We assume that the in-distribution (\id) samples belong to one of the $K$ known classes, and club all \ood samples into a new class called a \emph{reject}/\ood class. We do not attempt to identify which specific class (unseen label) the unknown sample belongs to. Our goal is to build a classifier to accurately detect \ood samples as the $(K+1)^\text{th}$ \ood class, with an objective to reject samples belonging to any novel class. 

Most techniques for \ood detection assume the availability of validation samples from the \ood set for tuning model hyper-parameters \cite{liang2018enhancing,lee2018simple,chen2021atom,hendrycks2018deep}. Based on the samples, the techniques either update the model weights so as to predict lower scores for the \ood samples, or try to learn correlation between activations and the output score vector \cite{lee2018simple}. Such approaches have limited utility as in most practical scenarios, either the \ood samples are not available, or cover a tiny fraction of \ood sample space. Yet, other class of techniques learn the threshold on the uncertainty of the output score using deep ensembling \cite{lakshminarayanan2017simple} or MC dropout \cite{gal2016dropout}. Understandably, \ood detection capability of these techniques suffer when the samples from a different \ood domain are presented. 

The other popular class of \ood detectors do not use representative samples from \ood domain, but generate them synthetically \cite{hendrycks2018benchmarking,Mohseni_Pitale_Yadawa_Wang_2020,neal2018open}. The synthetic samples can be used to train any of the earlier mentioned \sota models in lieu of the real \ood samples. This obviates the need for any domain specific \ood validation set. Such methods typically use natural corruptions (e.g. blur, noise, and geometric transformations etc.) or adversarial perturbations to generate samples near decision boundary of a classifier. This class  also have limited accuracy on real \ood datasets, as the synthetic images generated in such a way are visually  similar/semantically similar to the \id samples, and the behavior of a \dnn when shown natural \ood images much farther (in terms of $\ell_2$ distance in RGB space) from the \id samples still remains unknown. 



Recent theoretical works towards estimating or minimizing open set loss recommend training with \ood samples covering as much of the probable input space as possible. 
For example, \cite{xuji} show that a piece-wise \dnn model shatters the input space into a polyhedral complex, and prove that empirical risk of a \dnn model in a region of input space scales inversely with the density of training samples lying inside the polytope corresponding to the region. Similarly, \cite{fang2021learning} show that under an unknown \ood distribution, the best way to minimize the open set loss is by choosing \ood samples uniformly from the support set in the input space. Encouraged by such theoretical results, we propose a data augmentation technique which does not focus on generating samples visually similar to the \id samples but synthesizing \ood samples in two key regions of the input space: (i) finely distributed at the boundary of \id classes, and (ii) coarsely distributed in the inter-\id sample space (See Sec. \ref{subsec:polyhedralDecomp} for details). We list the key contributions:
\begin{enumerate}
\item We propose a novel data augmentation strategy, \textbf{C}ompou\textbf{n}ded \textbf{C}orruptions (CnC) for \ood detection. Unlike contemporary techniques \cite{guo2017calibration,hendrycks2018deep,lee2018simple,liang2018enhancing} the proposed approach does not need a separate \ood train or validation dataset.  
\item Unlike \sota techniques which detect \ood samples by lowering the confidence of \id classes \cite{bendale2015open,hendrycks2016baseline,lee2018simple,liu2020energy}, we classify \ood samples into a separate reject class. We show empirically that our approach  leads to clearer separation between \id and \ood samples in the embedding space (Fig. \ref{fig:K+1vsK}).
\item Our method does not require any input pre-processing at the test time, or a second forward pass with perturbation/noise. This makes it significantly faster in inference as compared to the other \sota methods \cite{hsu2020generalized,liang2018enhancing}.
\item Visualization and analysis of our results indicate that finer granularity of the polyhedral complex around the \id regions learnt by a model is a good indicator of performance of a \ood data augmentation technique. Based on our analysis, we also recommend higher entropy and diversity of generated \ood samples as good predictors for \ood detection performance. 

\item CnC yields \sota results on multiple benchmark datasets. Using DenseNet~\cite{huang2018densely}, CIFAR-100 as \id dataset, and SVHN~\cite{svhn} as \ood, we achieve TNR@TPR95 of $98.7\%$, AUROC of $99.7\%$ and Detection Error of $2\%$. This outperforms GODIN~\cite{hsu2020generalized} by a large margin of $18.4\%$ in TNR@TPR95 (Tab. \ref{tab:sota_comparison}).
\end{enumerate}

\section{Related Work}
\label{sec:relatedWorks}

Our approach is a hyper-parameter-free \ood detection technique, which does not need access to a validation \ood dataset. We review contemporary works below.

\paragraph{Hyper-parameter tuning using \ood data} 
This class comprises of \ood detection methods that fine-tune hyper-parameters on a validation set. ODIN~\cite{liang2018enhancing} utilizes temperature scaling with input perturbations using the \ood validation dataset to tune hyper-parameters for calibrating the neural networks. However, hyper-parameters tuned with one \ood dataset may not generalize to other datasets. Lee \etal \cite{lee2018simple} propose training a logistic regression detector on the Mahalanobis distance vectors calculated between test images' feature representations and class conditional Gaussian distribution at each layer.

\paragraph{Retraining a model using \ood data} 
G-ODIN \cite{hsu2020generalized} decompose confidence score along with modified input pre-processing for detecting \ood, whereas ATOM \cite{chen2021atom} essentially makes a model robust to the small perturbations, and hard negative mining for \ood samples.  MOOD \cite{lin2021mood} introduce multi-level \ood detection based on the complexity of input data, and exploit simpler classifier for faster \ood inference. 

\paragraph{Using a pre-trained model's score for \ood detection} 
Hendrycks and Gimpel \cite{hendrycks2016baseline} use maximum confidence scores from a softmax output to detect \ood. Liu \etal \cite{liu2020energy} use energy as a scoring function for \ood detection without tuning hyper-parameters. Shastry and Oore \cite{sastry2020detecting} leverage $p^\text{th}$-order Gram matrices to identify anomalies between activity patterns and the predicted class. Blundell \etal \cite{bendale2015open} focus on a closed world assumption which forces a \dnn to choose from one of the \id classes, even for the \ood data. \emph{OpenMax} estimates the probability of an input being from an unknown class using a Weibull distribution. 
G-OpenMax\cite{GenOpenMax} explicitly model \ood samples and report findings on small datasets like MNIST.

\paragraph{\ood detection using uncertainty estimation}  
\ood samples can be rejected by thresholding on the uncertainty measure. Graves \etal\cite{NIPS2011_4329}, Wen \etal\cite{wen2018flipout} propose anomaly detection based on stochastic Bayesian inference. Gal \etal \cite{gal2016dropout} propose  MC-dropout to measure uncertainty of a model using multiple inferences. Deep Ensembles \cite{lakshminarayanan2017simple} use multiple networks trained independently to improve uncertainty estimation. 

\paragraph{Data augmentation for \ood detection} 
This line of research augments the training set to improve \ood detection. Data augmentations like flipping and cropping generate samples that can be easily classified by a pre-trained classifier. Generative techniques based on VAEs, and GANs try to synthesize data samples near the decision boundary \cite{vos,lee2017training,li2018anomaly,perera2019ocgan,xiao2020likelihood,wang2017safer,anomaly}. Other data augmentation strategies do not directly target \ood detection, but domain generalization: SaliencyMix~\cite{uddin2021saliencymix}, CutOut\cite{devries2017improved}, GridMask\cite{chen2020gridmask}, AugMix \cite{hendrycks*2020augmix}, RandomErase \cite{zhong2017random}, PuzzleMix \cite{kim2020puzzle}, RandAugment \cite{RandAugment}, SuperMix \cite{dabouei2020supermix}. Mixup~\cite{zhang2017mixup} generates new data through convex combination of training samples and labels to improve DNN generalization. CutMix~\cite{yun2019cutmix} which generates samples by replacing an image region with a patch from another training image. The approach is not directly suitable for \ood detection, as the generated samples lie on the line joining the training samples, and may not cover the large input space\cite{xuji,fang2021learning}.  

\section{Proposed Approach}
\label{sec:cnc}

\begin{figure}
		\centering
		\includegraphics[width=\linewidth]{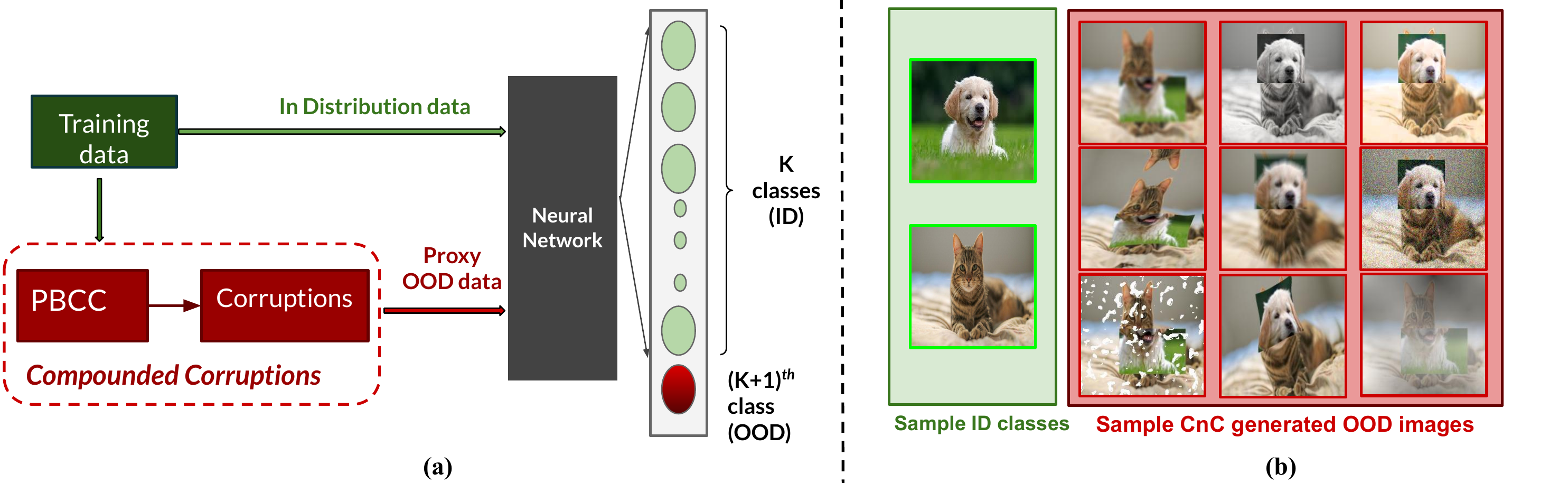}
		\caption{Creating augmented data samples using Compounded Corruptions (CnC). Pane (a) shows block diagram of the training procedure: first we take a patch based convex combination (PBCC) of patches chosen from image pair belonging to $K \choose 2$ labels; second, we apply corruptions on the data points obtained using PBCC. This proxy \ood data is then used to train a $(K+1)$ way classifier, where, first $K$ classes correspond to the \id classes and $(K+1)^{th}$ class contains synthesized \ood samples corresponding to reject/\ood class. Pane (b) shows CnC synthesized sample images from \texttt{cat} and \texttt{dog} classes. Intuitively, CnC gives two knobs for generating \ood samples: a coarse exploration ability through linear combination of two \id classes achieved through PBCC operation, and a finer warping capability through corruption of these images. The order of the two operations (PBCC before corruption) is important, as we show later.}
\label{fig:cncTraining}
\end{figure}

\subsection{Problem Formulation} 

We consider a training set, $\D_\text{in}^\text{train}$, consisting of $N$ training samples: $(x_n, y_n)^N_{n=1}$, where samples are drawn independently from a probability distribution: $\cP_{X,Y}$. Here, $X\in\X$ is a random variable defined in the image space, and $Y\in\Y = \{1,\ldots,K\}$ represents its label. Traditionally, a classifier $f_{\theta}:\mathcal{X} \rightarrow \mathcal{Y}$ is trained on in-distribution samples drawn from a marginal distribution $\cP_X$ of $X$ derived from the joint distribution $\cP_{X,Y}$. Let $\theta$ refers to model parameters and $\cQ_X$ be another distinct data distribution defined on the image space $\X$. During testing phase, input images are drawn from a conditional mixture distribution $\cM_{X | Z}$ where $Z \in \{0,1\}$, such that $\cM_{X \mid Z=0} = \cP_X$, and $\cM_{X \mid Z=1} = \cQ_X$. We define all $\cQ_X \nsim \cP_X$ as \ood distributions, and $Z$ is a latent (binary) variable to denote \id if $Z=0$ and \ood if $Z=1$.

One possible approach to detecting an \ood sample is if confidence of $f_\theta$ for a given input is low for all elements of $\mathcal{Y}$. However, we use an alternative approach where we learn to map \ood samples generated using our technique to an additional label $(K+1)$. Given any two \id samples $x_1,x_2 \sim \mathcal{P}_X$, we generate the synthetic data using the CnC operation $C(x_1,x_2):\X \times \X \rightarrow \X$. We then define an extended label set $\mathcal{Y}^+=\{1,\ldots,K+1\}$, and train a classifier $f_\theta^+$ over $\mathcal{Y}^+$. The goal is to train $f_\theta^+$ to implicitly build an estimate $\hat{Z}$ of $Z$, such that the output of $f_\theta^+$ is $(K+1)$ if $\hat{Z}=1$, and one of the elements of $\mathcal{Y}$ if $\hat{Z}=0$. 

\begin{figure}
	\centering
	\includegraphics[width=\linewidth]{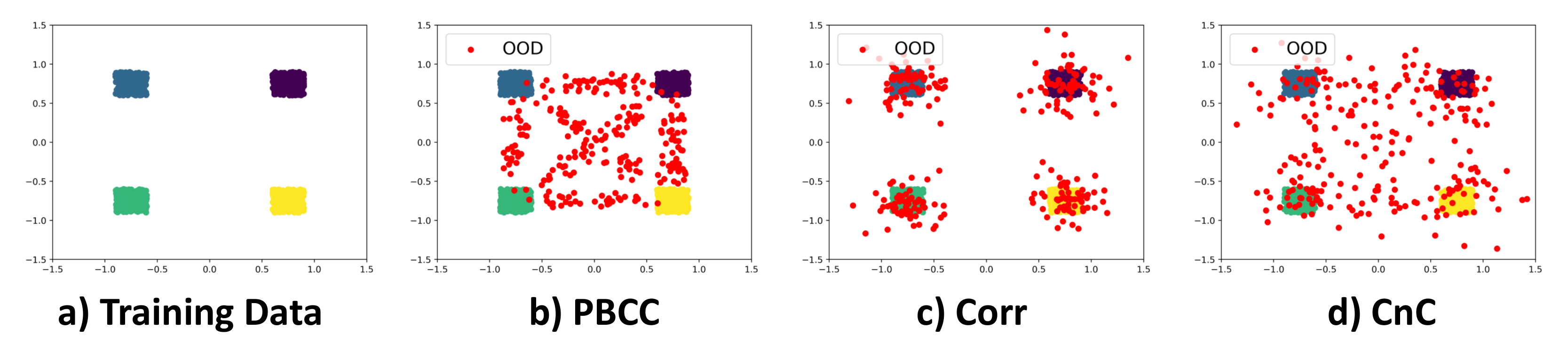}
	\caption{Intuition with an illustrative plot of \ood synthesis on a toy dataset with four \id classes. Each sample is in $\mathbb{R}^{2}$. Consider $\textbf{p}_1=(x_1,y_1)$, and  $\textbf{p}_2=(x_2,y_2)$ to be the two input samples belonging to distinct classes 1 and 2, then $\textbf{p}_3=(x_3,y_3)$ is the geometric convex combination of $\textbf{p}_1$ and $\textbf{p}_2$ such that: $\textbf{p}_3= \lambda \textbf{p}_1+(1 - \lambda)\textbf{p}_2$ , $0 \le \lambda \le 1$. 
	(a) training data corresponding to 4 distinct classes; Synthesised \ood points are in red; (b) PBCC generates \ood points through a convex combination of ID points from different classes in $4 \choose 2$ ways, whereas corruptions depicted in (c) can generate \ood points around each cluster. Observe that points generated by CnC spans wider \ood space including inter-ID-cluster area and outside the convex hull of \id points.} 
	\label{fig:umap}
\end{figure}

\subsection{Synthetic OOD Data Generation}
\label{subsec:synthDataGen}

Our synthetic sample generation strategy consists of following two steps. 

\paragraph{Step 1: Patch Based Convex Combination (PBCC)}
We generate synthetic samples by convex combination of two input images. Let $x \in \mathbb{R}^{W \times H \times C}$, and $y$ denote a training image and its label respectively. Here, $W, H, C$ denote width, height, channels of the image respectively. A new sample, $\xt$, is generated  by a convex combination of two training samples $(x_{A},y_{A})$, and $(x_{B},y_{B})$:
\begin{equation}
	\Tilde{x} = \mathbf{M} \odot x_{A} + (\mathbf{1}  - \mathbf{M}) \odot x_{B}.
\end{equation}
Here, $x_{A}$ and $x_{B}$ do not belong to a same class ($y_A \ne y_B$), and $\mathbf{M} \in \{0,1\}^{W \times H}$ denotes a rectangular binary mask that indicates which region to drop, or use from the two images. $\mathbf{1}$ is a binary mask filled with ones, and $\odot$ is element-wise multiplication. To sample $\mathbf{M}$, we first sample the bounding box coordinates $\mathbf{B} = (r_x, r_y, r_w, r_h)$, indicating the top-left coordinates, and width, and height of the box. The region $\mathbf{B}$ in $x_{A}$ is cut-out and filled in with the patch cropped from $\mathbf{B}$ of $x_{B}$. The coordinates of  $\mathbf{B}$ is uniformly sampled according to: $r_{x} \sim \text{U}(0, W), r_{w} = W\sqrt{1 - \lambda}$ and similarly, $r_{y} \sim \text{U}(0, H), r_{h} = H\sqrt{1 - \lambda}$. Here, $\lambda \in [0,1]$ denotes the crop area ratio, and is fixed at different values for generating random samples. The cropping mask $\mathbf{M}$ is generated by filling zeros within the bounding box $\mathbf{B}$ and ones outside. We generate the samples by choosing each pair of labels in $K \choose 2$ ways, and then randomly selecting input images corresponding to the chosen labels. This generates \ood samples spread across various inter-class regions in the embedding space. For ablation on range of $\lambda$ to ensure that a large number of \ood samples are generated outside the ID clusters see supplementary.  We label all generated samples as that of the $(K+1)^\text{th}$ reject class.

PBCC and CutMix \cite{yun2019cutmix}: Note that PBCC and CutMix\cite{yun2019cutmix} both rely on the same basic operation \textbf{convex combination of images}, but for two very different objectives. Whereas, CutMix uses the combination step to guide a model to attend on less discriminative parts of objects e.g. leg as opposed to head of a person letting the network generalize better on object detection. On the other hand, we use PBCC as a first step for \ood data generation, where the operation generates samples in a large \ood space between a pair of classes in $K \choose 2$ ways. 

PBCC Shortcomings: Note that PBCC performs a convex combination of the two \id images belonging to two distinct classes. Hence, unlike adversarial perturbations, it is able to generate sample points far from the \id points in the RGB space. However, still it can generate samples from only within the convex hull of the \id points corresponding to all classes.Thus, as we show in our ablation studies, sample generated using this step alone are insufficient to train a good \ood detector. Below we show how to improve upon the shortcoming of PBCC.

\paragraph{Step 2: Compounded Corruptions} We aim to address the above shortcomings by using corruptions on top of PBCC generated samples, thus increasing the sample density in inter-class regions as well as generating samples outside the convex hull. We reason that such compounded corruptions increase the spread of the augmented data to a much wider region. Thus, a reasoning based on ``per sample" generalisation error bound from \cite{xuji}:[Fig. 1, Equation 11] could be utilized for our problem. \cite{xuji} constructs an input-dependent generalization error bound by analysing the subfunction membership of each input, and show that generalisation error bound improves with smoother training sample density (as defined by number of samples in each region). Intuitively, corruptions over PBCC produces a smoother approximation of \id classes with a finer fit at the \id class boundary. A detailed analysis is given in Fig. \ref{subsec:polyhedralDecomp}. To give an intuitive understanding, Fig \ref{fig:umap} shows visualizations of the generated \ood samples in red using a 4 class toy dataset in two dimensions.

Hendrycks \etal ~\cite{hendrycks2018benchmarking} benchmark robustness of a \dnn using $15$  algorithmically generated image corruptions that mimic natural corruptions. Each corruption severity ranges from $1$ to $5$ based on the intensity of corruption, where $5$ is most severe. The corruptions can be seen as perturbing a sample point in its local neighborhood, while remaining in the support space of the probability distribution of valid images. We apply these corruptions on the samples generated using PBCC step described earlier. Together, PBCC, and corruptions, allow us to generate a synthetic sample far from, and outside the convex hull of \id samples. At the same time, unlike pure random noise images, the process maintains plausibility of the generated samples. Specifically we apply following corruptions: Gaussian noise, Snow, Fog, Contrast, Shot noise/Poisson noise, Elastic transform, JPEG compression, and blur such as Defocus, Motion etc. 

Fig. \ref{fig:cncTraining} gives a pictorial overview of the overall proposed scheme with a few \ood image samples generated by our approach. CnC formulates the problem as $(K+1)$ class classification which improves the model representation of underlying distribution, and at the same time improves \dnn calibration as seen in Sec. \ref{subsec:otherBenefitsCnC}. Please see Suppl.  for the precise steps of our algorithm. 


\begin{figure}
\centering
\includegraphics[width=\linewidth]{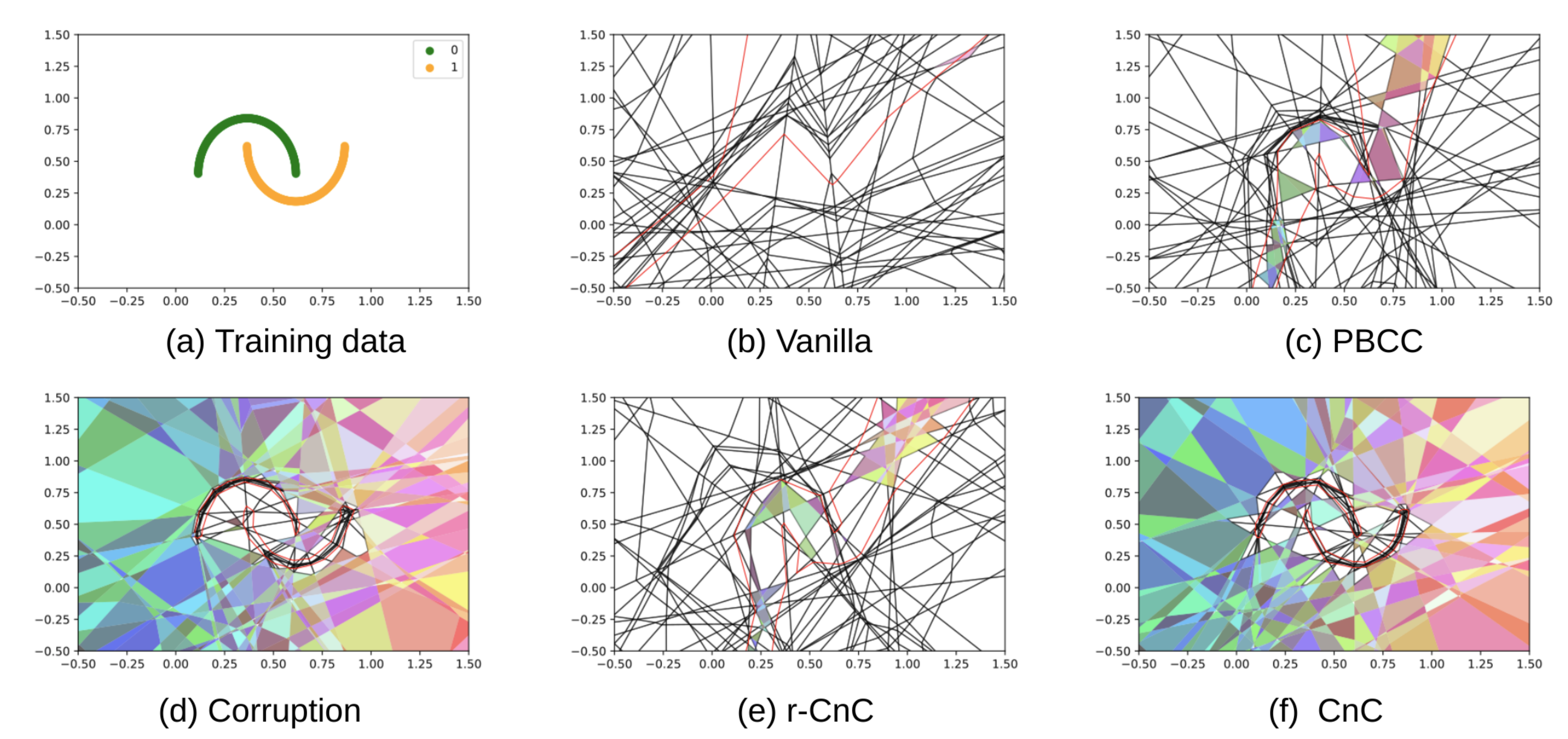}
\caption{Visualization of trained classifiers as a result of \ood augmentation. A ReLU type \dnn is trained on the two-dimensional half-moon data set shown in (a). The shattered neural networks \cite{hein2019relu} show that CnC has the tightest fit around the ID regions, as measured by the area of the (white colored) polytopes in which no training \id point is observed but a network predicts a point in that region as \id. The measured areas for such polytopes are (b)\textbf{Vanilla training without data augmentation}:5.65,  (c)\textbf{PBCC}: 8.20, (d) \textbf{Corruption}:0.40,  (e)\textbf{r-CnC}: 5.66, (f)\textbf{CnC}: 0.37. Note: \cite{xuji} state that the more densely supported a polytope is by the training set, the more reliable the network is in that region. Hence, the samples declared \id in the regions where no \id sample is observed may actually be \ood with high probability. We observe that PBCC/r-CnC/Vanilla, all predict \id in many such polytopes. Note: r-CnC we reverse the order of PBCC and corruptions \textbf{[Best viewed at 200\%]}} 
\label{fig:polytope}
\end{figure}

\subsection{CnC Analysis via Polyhedral Decomposition of Input Space}
\label{subsec:polyhedralDecomp}


While we validate the improved performance of CnC in Sec. \ref{sec:expnresults}, in this section we seek to provide a plausible explanation for the CnC's performance.  
We draw inspiration from theoretical support provided in recent work by \cite{xuji} who formally derive and empirically test prediction unreliability for ReLU based neural networks.

Consider a ReLU network with $n$ inputs and $m$ neurons in total. \cite{xuji} show that parameters of a trained model partition the input space into a polyhedral complex (PC) consisting of individual \textit{convex} polytopes (also called \textit{activation regions} in \cite{xuji}). See Fig. \ref{fig:polytope} for an example with a 2D input space. Each possible input corresponds to a unique state (active or inactive) of each of the $m$ ReLU neurons, and the interior of each polytope corresponds to a unique combination of states of all $m$ neurons. Thus a trained network behaves linearly in the interior of corresponding polytopes. Each edge in the PC corresponds to the state flip of a single neuron (active to inactive, or vice versa).

For the purpose of classification based on the final layer activation, a key corollary from \cite{xuji} is that \textit{the decision boundary between two classes must be a straight line within a polytope, and can only turn at the vertices}. This is an immediate consequence of the observation that the decision boundary is the locus along which the two highest activations (most probable labels) in the output layer remain equal to each other. This implies that smaller polytopes near the decision boundary are needed for finer control over the boundary between training samples from different classes. Note also that the authors in \cite{xuji}:[equation (11)] infer that (paraphrased) ``the more a subfunction (polytope) is surrounded by samples with accurate predictions, the lower its empirical error and bound on generalization gap, and thus the lower its expected error bound''. 

The key question from \ood detection perspective is, how do we force a network to create tighter polytopes at the \id class decision boundaries? We believe the answer is to distribute a large number of the augmented samples (over which we have control) with contrasting \ood and \id labels all around each \id region, forcing the decision boundary to form a tight bounding surface. At the same time, we must also retain a good fraction of the augmented samples in the open space between ID classes, which can be covered by relatively large polytopes (recall that the maximum number of polytopes is bounded by the number of neurons, and thus small polytopes in one region may need to be traded off by larger polytopes in another region). Neglecting the inter-\id space entirely would run the risk of creating very large polytopes in this region, which increases the empirical error bound (\cite{xuji}:[equation (5) and (11), large subfunctions have low probability mass and hence higher error bound. Refer Supplementary for further details.]. CnC lets us achieve this dual objective by using compounding to sample the space between \id classes, and corruption to pepper the immediate neighborhoods around \id classes (especially for $\lambda$ values near $0$ and $1$).


In Fig. \ref{fig:polytope}, we show polyhedral complex corresponding to the \dnn models trained on two-dimensional half-moon dataset \cite{hein2019relu,jordan2019provable}, and \ood samples generated using various techniques. The first plot shows the input space with training samples from two \id classes (green and yellow semicircles). The learnt polytope structure for vanilla uses a neural network of size $[2,32,32,2]$, while the remaining three plots use $[2,32,32,3]$ (with an additional \textit{reject}/\ood class).   

Recall from Fig. \ref{fig:umap} that PBCC produces samples sparsely between the \id classes, but not around the \id class boundaries. Pure corruptions produce samples only near and on \id classes, but not in the inter-\id space. On the other hand, CnC produces samples both near the \id boundaries as well as in the inter-\id space. In Fig. \ref{fig:polytope}, we define any polytope that is fully or partially (decision boundary crosses through it) classified as \id, as an ``\id classified polytope'' and mark it in white color. \textit{Visually, we can see that the white polytopes occupy a smaller total area when we compare Vanilla to CnC, with the actual values noted in the caption. This indicates that the CnC produces the tightest approximation of \id classes in our example, which in turn leads to better \ood detection.} Though we show for two-dimensional data, we posit that the same generalizes to higher dimensional input data as well, and is the reason for success of CnC based \ood detection.

CnC and Robustness to Adversarial Attacks: Note that, small polytopes in the input space partitioned by a \dnn may also provide better safety against black box adversarial attacks as suggested by \cite{hein2019relu,jordan2019provable}. This is because the black box adversarial attacks extrapolate the gradients based upon a particular test sample. Since the linearity of the output, and thus the gradients is only valid inside a polytope, smaller polytopes near the \id or in the \ood region makes it difficult for an adversary to extrapolate an output to a large region. However, since adversarial robustness is not the focus of this paper, we do not further explore this direction.

\subsection{Training Procedure}
\label{sec:trainingsec}

We train a $(K+1)$ class classifier network $f_{\theta}^+$, where first $K$ classes correspond to the multi-classification ID classes, and the $(K+1)^{th}$ class label indicates the \ood class. 
Our training objective takes the form: 
\begin{align}
	\mathcal{L} = \minimize_{\theta} 
	\hspace{0.35cm} 
	\mathbb{E}_{(x, y) \sim  D_\text{in}^\text{train}} [\lce(x, y;f_{\theta}^+(x))] \nonumber\\
	+ \alpha \cdot \mathbb{E}_{(x, y) \sim D_{pbcc}^{corr}}[\lce(x, K+1; f_{\theta}^+(x))],
\end{align}
where $\lce$ is the cross entropy loss, $f_{\theta}^+(x)$ denotes the softmax output of neural network for an input sample $x$. We use $\alpha = 1$ in our experiments based on the ablation study reported in the supplementary material. For above experiments setup we set the ratio of \iid:\ood training points as $1:1$.

\subsection{Inference}
\label{subsec:inference}

After training, we obtain a trained model $F^+$. We use $F^+(x)[K+1]$ as the \ood score of $x$ during testing, and define an \ood detector $D(x)$ as:
\begin{equation}
	D(x) =
	\begin{cases}
		0, & \text{if}\ F^+(x)[K+1] > \delta \\
		1, & \text{if}\ F^+(x)[K+1] \leq \delta \\
	\end{cases}
	\label{equ:ood_score}
\end{equation}
where, $D(x) = 0$ indicates an \ood prediction, and $D(x) = 1$ implies an \id sample prediction. $\delta$ is a threshold 
such that TPR, i.e., fraction of \id images correctly classified as \id is $95\%$. For images which are characterized as \id by $D(x)$, the labels are given as:. 
\begin{equation}
	\hat{y} = \argmax_{i \in 1, \ldots, K} F^+(x)_{i}
\end{equation}

\section{Dataset and Evaluation Methodology}
\label{sec:datasets_and_evaluation}


In-Distribution Datasets: For \id samples, we use SVHN (10 classes) \cite{svhn}, CIFAR-10 (10 classes), CIFAR-100 (100 classes)\cite{Krizhevsky2009LearningML} containing images of size $32\times32$. We also use TinyImageNet (200 classes) \cite{le2015tiny} containing images of resolution $64\times64$ images. Out-of-Distribution Datasets: For comparison, we use the following \ood datasets: TinyImageNet-crop (TINc), TinyImageNet-resize (TINr), LSUN-crop (LSUNc), LSUN-resize (LSUNr), iSUN, SVHN. Evaluation Metrics: We compare the performance of various approaches using TNR@TPR95, AUROC and Detection Error. See Suppl. for description on evaluation metrics.

\begin{table}
\small
\begin{center}
\begin{tabular}{c|l|ccccc}
\hline
	\centering
\multirow{2}{*}{$\mathcal{D}_\text{in}^\text{train}$} & 
\multirow{2}{*}{\textbf{\scriptsize{Method}}} & \textbf{\scriptsize{TNR@TPR95}} & \textbf{\scriptsize{AUROC}} & \textbf{\scriptsize{DetErr}} & \textbf{\scriptsize{ID Acc.}}
\\ 
& & $\pmb{\uparrow}$ & $\pmb{\uparrow}$ & $\pmb{\downarrow}$ & $\pmb{\uparrow}$
\\ 
\hline
\multirow{7}{*}{\vertext{CIFAR-10} \vertext{DenseNet-BC}} 
& MSP (ICLR'17) \cite{hendrycks2016baseline}  & 56.1 & 93.5 & 12.3 & 95.3 \\
& ODIN (ICLR'18)\cite{liang2018enhancing}  & 92.4 & 98.4 & 5.8 & 95.3 \\
& Maha(NeurIPS'18)\cite{lee2018simple} & 83.9 & 93.5 & 10.2 & 95.3  \\
& Gen-ODIN (CVPR'20)\cite{hsu2020generalized} & 94.0 & 98.8 & 5.4 & 94.1 \\
& Gram Matrices(ICML'20)\cite{sastry2020detecting} & 96.4  & 99.3 & 3.6 & 95.3\\
& ATOM(ECML'21) \cite{chen2021atom} & 98.3 & 99.2 & 1.2 & 94.5 \\
&\textbf{CnC(Proposed)} & \textbf{98.4 $\pm$ 0.8} & \textbf{99.5 $\pm$ 1.2} & \textbf{2.7 $\pm$ 0.2} & 94.7
\\
\hline
\multirow{7}{*}{\vertext{CIFAR-100} \vertext{DenseNet-BC}} 
& MSP (ICLR'17) \cite{hendrycks2016baseline}  & 21.7 & 75.2 & 31.4 & 77.8\\
& ODIN (ICLR'18)\cite{liang2018enhancing} & 61.7 & 90.6 & 16.7 & 77.8 \\
& Gen-ODIN (CVPR'20)\cite{hsu2020generalized} & 86.5 & 97.4 & 8.0 & 74.6  \\
& Maha (NeurIPS'18)\cite{lee2018simple}& 68.3 & 92.8 & 13.4 & 77.8  \\
& Gram Matrices(ICML'20)\cite{sastry2020detecting}& 88.8  & 97.3 & 7.3 & 77.8 \\
& ATOM(ECML'21)\cite{chen2021atom} & 67.7 & 93  & 5.6  & 75.9\\
& \textbf{CnC(Proposed)} & \textbf{97.1 $\pm$ 1.4} & \textbf{98.5 $\pm$ 0.4} & \textbf{4.6 $\pm$ 0.6} & 76.8
\\
\hline
\multirow{3}{*}{\vertext{TIN} \vertext{RN50}}
& MSP (ICLR'17) \cite{hendrycks2016baseline}  & 53.15 & 85.3 & 22.1 & 57.0\\
& ODIN (ICLR'18)\cite{liang2018enhancing}  & 68.5 & 93.7 & 12.3 & 57.0\\
& \textbf{CnC(Proposed)} & \textbf{97.8 $\pm$ 0.8} & \textbf{99.6 $\pm$ 0.2} & \textbf{2.1 $\pm$ 0.2} & 60.5\\
\hline
\multirow{3}{*}{\vertext{C-10} \vertext{WRN}}
& OE (ICLR'19) \cite{hendrycks2018deep} & 93.23 & 98.64 & 5.32 &   94.8 \\
& EBO (NeurIPS'20)\cite{liu2020energy} & \textbf{96.7} & {99.0} & \textbf{3.83} &  95.2 \\
& \textbf{CnC(Proposed)} & {96.2 $\pm$ 1.5} & \textbf{99.02 $\pm$ 0.1} & { 4.5 $\pm$ 0.8} & 94.3 \\
\hline
\multirow{3}{*}{\vertext{C-100} \vertext{WRN}}
& OE (ICLR'19) \cite{hendrycks2018deep} & 47.35 & 86.02 & 21.24 &  75.6 \\
& EBO (NeurIPS'20)\cite{liu2020energy} & 54.0 & 86.65 & 19.7 &   75.7 \\
&\textbf{CnC(Proposed)} & \textbf{97.6 $\pm$ 0.9} & \textbf{99.5 $\pm$ 0.1} & \textbf{ 2.2 $\pm$ 0.3} & 75.1\\
\hline 
\end{tabular}%
\end{center}

\caption{\small Comparison of competing \ood detectors. TIN: TinyImageNet, and RN50: ResNet50, WRN : WideResNet-40-2  Values are averaged over all \ood benchmark datasets. We give individual dataset-wise results in the supplementary. Note that ATOM\cite{chen2021atom}, and OE \cite{hendrycks2018deep} require large image datasets like 80-Million Tiny Images \cite{80tinyimages} as representative of \ood samples. However, CnC synthesises its own \ood dataset using the \id training data. 
CnC models were trained using the same configuration as defined by OE \cite{hendrycks2018deep} and EBO \cite{liu2020energy} paper, with the exception that CnC did not use any external auxiliary OOD dataset like \cite{80tinyimages} in training.
CnC reasults are averaged on 3 evaluation runs.}

\label{tab:sota_comparison}
\end{table}

\begin{table}
	\centering
	\small
		\begin{tabular}{l|ccc}
			\toprule[1pt]
			\textbf{Data Augmentation Methods} & \textbf{\scriptsize{TNR (95\% TPR)}} & \textbf{\scriptsize{AUROC}} & \textbf{\scriptsize{Detection Err}}\\ 
			& $\pmb{\uparrow}$ & $\pmb{\uparrow}$ & $\pmb{\downarrow}$
\\ 
			\cline{1-4}
			Mixup (ICLR'18) \cite{zhang2017mixup}  & 60.6 & 90.9 & 15.5  \\
			CutOut (arXiV'17) \cite{devries2017improved} & 80.8 & 94.8 & 10  \\
			CutMix (ICCV'19) \cite{yun2019cutmix} & 83.2 & 92.7 & 8.6    \\
			GridMask (arXiV'20) \cite{chen2020gridmask} & 50.3 & 79.1 & 23.6  \\
			SaliencyMix (ICLR'21) \cite{uddin2021saliencymix} & 85.3  & 95.7 & 8.0 \\
			AugMix (ICLR'20) \cite{hendrycks*2020augmix} & 81.3 & 94.6 & 11.2 \\
			RandomErase (AAAI'20) \cite{zhong2017random} & 41.9 & 68.1 & 24.2 \\
			Corruptions (ICLR'19) \cite{hendrycks2018benchmarking} & 98.0 & 99.4 & 2.8 \\
			PuzzleMix (ICML'20) \cite{kim2020puzzle} & 66.8 & 84.1 & 15.2 \\
			RandAugment (NeurIPS'20) \cite{RandAugment} & 89.5 & 97.9 & 4.7 \\
			Fmix (ICLR'21) \cite{harris2021fmix} & 73 & 90.3 & 12.6\\
			Standard Gaussian Noise & 71.5 & 93.2 & 11.7 \\
			\textbf{CnC(Proposed)} & \textbf{98.4 $\pm$ 0.8} & \textbf{99.5 $\pm$ 1.2} & \textbf{2.7 $\pm$ 0.2} \\
			\bottomrule[1pt]
		\end{tabular}
	\caption{\small Comparison with other synthetic data generation methods. We consider CIFAR10 as \id. The values are averaged over all OOD benchmarks. We have used DenseNet\cite{huang2018densely} as the architecture for all methods trained for $(K+1)$ class classification. Samples obtained through the listed data augmentation schemes were assumed to be of $(K+1)^\text{th}$ class. Observe that CnC has superior \ood detection performance. We report average and standard deviation of CnC trained models computed over 3 runs.}
	
	\label{tab:dataAug}
\end{table}

\section{Experiments and Results}
\label{sec:expnresults}

To show that our data augmentation is effective across different feature extractors, we train using both DenseNet-BC~\cite{huang2018densely} and ResNet-34~\cite{he2016identity}. DenseNet has 100 layers with growth rate of 12. WideResNet \cite{zagoruyko2016wide} models have the same training configuration as  \cite{liu2020energy}. 

\subsection{Comparison with State-of-the-art}

\paragraph{\ood Detection Performance: }
Tab. \ref{tab:sota_comparison} shows comparison of CnC with recent state-of-the-art. The numbers indicate averaged \ood detection performance on 6 datasets as mentioned in Sec. \ref{sec:datasets_and_evaluation} (TinyImagenet, TinyImageNet-crop (TINc), TinyImageNet-resize (TINr), LSUN-crop (LSUNc), LSUN-resize (LSUNr), iSUN, SVHN) 
with more details included in the supplementary.
We would like to emphasize that CnC does not need any validation \ood data for fine-tuning. But ODIN~\cite{liang2018enhancing} and Mahalanobis ~\cite{lee2018simple} require \ood data for fine-tuning the hyper-parameters; the hyper-parameters for ODIN and Mahalanobis methods \cite{liang2018enhancing,lee2018simple}  are set by validating on 1K images randomly sampled from the test set $\mathcal{D}^\text{test}_\text{in}$. Tab. \ref{tab:sota_comparison} clearly shows that CnC outperforms the existing methods.

\paragraph{Comparison with Other Data Generation Methods}: Tab. \ref{tab:dataAug} shows how CnC fairs against recent \ood data generation methods. In each case we train a $(K+1)$ way classier where first \emph{K} classes correspond to \id and $(K+1)^\text{th}$ class comprised of \ood data generated by corresponding method.  As seen from the table, CnC outperforms the recent data augmentation schemes. 

\subsection{Other Benefits of CnC}
\label{subsec:otherBenefitsCnC}

\paragraph{Detecting Domain Shift as \ood:}

\begin{table}
\centering
\small
		\begin{tabular}{l|ccc}
			\toprule[1pt]
			\textbf{Method} & \textbf{TNR@0.95TPR} & \textbf{AUROC} & \textbf{DetErr} \\ 
			\hline
			MSP (ICLR'17) \cite{hendrycks2016baseline}  & 24.4 & 80.1 & 26.5 \\
			ODIN (ICLR''18) \cite{liang2018enhancing}  & 46.0 & 88.6 & 18.9  \\
			Gen-ODIN (CVPR'20) \cite{hsu2020generalized} & 45.0 & 88.7 & 18.8  \\
			Mahalanobis (NeurIPS'18) \cite{lee2018simple} & 14.0 & 56.2 & 41.6  \\
			Gram Matrices (ICML'20) \cite{sastry2020detecting} & 35.0  & 81.5 & 25.8 \\
			\textbf{CnC (Proposed)} & \textbf{60.0} & \textbf{91.6} & \textbf{15.7}\\
			\bottomrule[1pt]
		\end{tabular}%
	\caption{\small Detecting domain shift using CnC. A model trained with CnC data  on CIFAR-100 as the \id using DenseNet-BC \cite{huang2018densely} feature extractor can successfully detect the domain shift when observing ImageNet-R at the test time.}
	\label{tab:domain_shift}
\end{table}

We analyze if a model trained with 
 CnC augmented data can detect non-semantic domain shift, i.e. images with the same label but different distribution. For the experiments we use a model trained using CIFAR-100 as \id, and ImageNet-O/ImageNet-R/Corrupted-ImageNet~\cite{hendrycks2019nae} as the \ood.
While testing, we downsample the images from ImageNet-O, ImageNet-R and TinyImageNet-C to a size of $32 \times 32$. Tab. \ref{tab:domain_shift} shows results on ImageNet-R OOD dataset. We outperform the next best technique by 14\% on TNR@0.95TPR, 2.9\% in AUROC, 3.1\% in detection error. See supplementary for results on ImageNet-O and Corrupted ImageNet. 


\paragraph{Model Calibration}
Another benefit of training with CnC is model calibration on \id data as well. A classifier is said to be calibrated if the confidence probabilities matches the empirical frequency of correctness \cite{guo2017calibration,hebbalaguppe2022stitch}, hence a crucial to measure of trust in classification models. Tables in the supplementary show the calibration error for a model trained on CIFAR-10, and CIFAR-100 as the \id data, with CnC samples as the $\text(K+1)^\text{th}$ class. 
Note that the calibration error is measured only for the \id test samples. We compare the error for a similar model, trained using only \id train data, and calibrated using temperature scaling (TS) \cite{guo2017calibration}. 

\paragraph{Time Efficiency}
For  applications  demanding  real-time  performance, it is crucial to have low latency in systems using \dnn for inference. Supplementary reports the competative performance of our method.

\subsection{Ablation Studies}


\emph{Rationale for Design choice of K vs. (K+1) Classifier}
%
We empirically verify having a separate class helps in better optimization/learning during training a model using CnC augmentation. Fig. \ref{fig:K+1vsK} shows the advantages of using a ($K+1$) way classifier as compared to standard $K$ class training with better \id-\ood separation. 
Supplementary material details the advantage of CnC with ACET \cite{hein2019relu} (CVPR'19) for uncertainty quantification on a half-moon dataset.

\paragraph{Recommendation for a Good \ood detector}

\begin{table}[t]
\centering
\small
	
	\begin{tabular}{l|ccccc}
		\specialrule{1pt}{0pt}{0pt}
		\multirow{3}{*}{\textbf{Method}} & \scriptsize{\textbf{TNR@}} & \multirow{2}{*}{\scriptsize{\textbf{AUROC}}} & \multirow{2}{*}{\scriptsize{\textbf{DetErr}}} & \scriptsize{\textbf{Mean}} & \scriptsize{\textbf{Mean}} \\ 
		&\scriptsize{\textbf{0.95TPR}} &  &  & \scriptsize{\textbf{Diversity}} & \scriptsize{\textbf{Entropy}} \\ 
		& $\pmb{\uparrow}$ & $\pmb{\uparrow}$ & $\pmb{\downarrow}$ & $\pmb{\uparrow}$ & $\pmb{\uparrow}$ \\ 
		\specialrule{0.75pt}{0pt}{0pt}
		PBCC  & 93.7 & 98.6 & 6.2 & 2.30 & 0.33 \\
		Corruptions  & 95.5 & 97.4 & 3.5 & 2.68 & 0.38 \\
		CnC & \textbf{98.3} & \textbf{99.6} & \textbf{2.6} & \textbf{3.40} & \textbf{0.80}
		\\
		\specialrule{1pt}{0pt}{0pt}
	\end{tabular}
	\caption{\small Using entropy/diversity of synthesized data to predict quality of \ood detection. Please refer to text for more details.}
	\label{tab:diversity_entropy}
\end{table}

We performed detailed comparison of various configurations of our technique to understand the quantitative scores which can predict the quality of an \ood detector.  For the experiment we keep the input images used same across configs, PBCC and corruptions applied are also fixed to remove any kind of randomness.
 We use ResNet34 as feature extractor for all methods. CIFAR-10 is used as ID dataset and TinyImageNet-crop as \ood dataset. We observe that the quality of \ood detection improves as the diversity, and entropy of the synthesized data increases (Tab \ref{tab:diversity_entropy}). Here, entropy is computed as the average entropy of the predicted probability vectors by the $K$ class model for the synthesized data. We adapt data diversity from Zhang \etal \cite{zhang2019active} to measure diversity of \ood data. Refer supplementary for  Algorithm for diversity computation.

\begin{figure}
	\begin{center}
		\includegraphics[width= \linewidth]{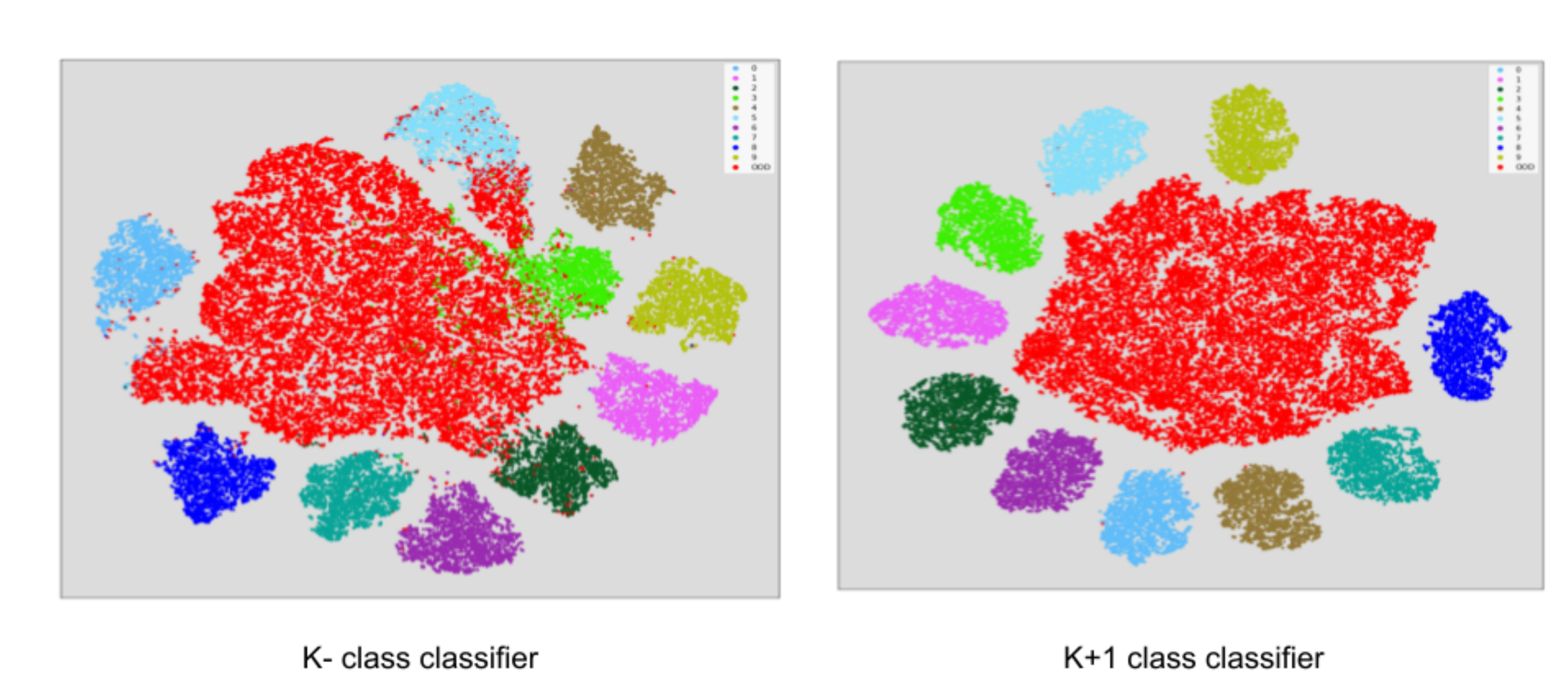}
		\caption{We show sample t-SNE plots for $K$ Vs. $(K+1)$ classifiers, where CIFAR-10 is used as ID and SVHN is used as \ood(marked in red). The K-class classifier uses temperature scaling (TS) \cite{guo2017calibration}, where T is tuned on SVHN test set. On the other hand, the $(K+1)$ class classifier uses SVHN data for $(K+1)^\text{th}$ class during training. The visualization shows that the \ood data (marked in \textcolor{red}{red}) is better separated in a $(K+1)$-class classifier as compared to a $K$-class classifier}
		\label{fig:K+1vsK}
	\end{center}
\end{figure}

\paragraph{Limitations of CnC data augmentation}: Introduction of additional synthetic data indeed increases training time. For eg., training a model with CnC data on TinyImageNet dataset takes 10 mins. 23 secs./epoch, whereas without CnC data it takes 5 mins 30 secs./epoch on the same Nvidia V100 GPU. Performance gain the overhead of training time can be discounted as inference time remains same. We assume the absence of adversarial intentions in this approach, Our method fails when tested against $L_{\infty}$ norm bounded perturbed image. In future we intend to look at \ood detection using CnC variants for non-visual domains. 
\section{Conclusions}
\label{sec:conclusions}

We have introduced \textbf{C}ompou\textbf{n}ded \textbf{C}orruptions(CnC), a novel data augmentation technique for \ood detection in image classifiers. CnC outperforms all the \sota \ood detectors on standard benchmark datasets tested upon. The major benefit of CnC over \sota is absence of \ood exposure requirement for training or validation. We also show additional results for robustness to distributional drift, and calibration for CnC trained models. CnC requires just one inference pass at the test time, and thus has much faster inference time compared to \sota. Finally, we also recommend high diversity and entropy of the synthesized data as good measures to predict quality of \ood detection using it.


\section{Acknowledgements}
Thanks to Subhashis Banerjee, Rahul Narain, Lokender Tiwari, Sandeep Subramanian for insightful  comments.

\title{Supplemental Material: \\A Novel Data Augmentation Technique for Out-of-Distribution Sample Detection using Compounded Corruptions}

\author{
	Ramya Hebbalaguppe $^{1,2}$ 
	\quad Soumya Suvra Ghosal$^{1}$ 
	\quad Jatin Prakash$^{1}$ 
	\quad Harshad Khadilkar$^{2}$
	\quad Chetan Arora$^1$ 
	\\ 
	$^1$Indian Institute of Technology Delhi, India
	\quad$^2${TCS Research}
	\\
}
%
\institute{}
%
\titlerunning{OOD sample detection using compounded corruptions}
%
%
%
%
\maketitle              

This supplementary material comprises of: 
\begin{enumerate}
\item Algorithm \ref{alg:cnc} gives the precise steps to generate CnC \ood samples
  	\item  See Sec. \ref{sec:ood_detect} for details of hyperparameters used for some of the competing methods in our comparisons.
  	\item  Sec. \ref{sec:trainDet} details training methodology and lists the evaluation metrics 
  	\item Rationale for choosing a $(K+1)$ class classifier over a $K$ class one: Refer Figure \ref{fig:K+1vsK} for the t-SNE visualisation that shows better \id-\ood separation. In Section \ref{sec:kplus1VsK} we study how choosing a $(K+1)$ class classifier over a $K$ class classifier reduces the number of false positives and false negatives (Refer Tables \ref{tab:confusion_matrix_cnc} and \ref{tab:confusion_matrix_msp}). Fig. \ref{fig:uncertainty} shows confidence plot of a NN trained using CnC data on data points far away from \id data. 

  	\item Ablation studies to determine choice of $\alpha$ used in our optimisation objective (Equation 2 in the main paper) such that it minimises the \ood detection error (Fig. \ref{fig:alpha})
  	
  	\item Tab. \ref{tab:summary_results_conceptual_comparison} gives summary performance/conceptual differences of various approaches.
  	
  	\item Comparison with other data synthesis configurations
  	
  	\item Time efficiency during inference.
  	\item More experiments on datasets with distributional shift. See Sec. \ref{sec:suplRob} where we study \ood detection performance on datasets such as Imagenet-O, Corrupted \ood, and Imagenet-R.
  	\item Efficacy of our method on similar looking classes as \ood. We show a sample illustration of performance of CIFAR~10 trained model which has the  \texttt{cat} class during training but tested on a very similar but different class images like \texttt{cougar}, \texttt{female lion}, and \texttt{tiger} downloaded from internet. See Figure \ref{fig:cougar} for further details.
  	\item More comparisons of CnC with state-of-the-art (SOTA) methods for other combinations of \id-\ood datasets apart from the tables reported in the main manuscript. To ensure that CnC performance is feature extractor agnostic, we consider two standard feature extractors, DenseNet-BC and Resnet-34. Please refer to Tables~\ref{tab:densenet10} and \ref{tab:densenet100} for results obtained using DenseNet as feature extractor. 
  	\item Comparison of ROC curves for CnC, and other data augmentation methods (see Figure \ref{fig:auroc}).
  	\item Additional results of comparison with other data generation methods: Table 3 in main manuscript gave an overview of comparison of CnC with other $12$ synthetic data generation methods when CIFAR10 is considered as ID and TinyImagenet(crop) as \ood. 
  	\item Limitations of our method: we acknowledge the dip in performance of CnC for non-semantic distribution shift, where the class labels remain the same but the distribution shifts. This is a typical setting for domain adaptation. However, we note that CnC still outperforms the recent state-of-the-art (see Table \ref{tab:cnc_lim}). 
\end{enumerate}


\begin{algorithm}
	\label{algo_Rasha}
	\SetKwInput{KwInput}{Inputs}                
	\SetKwInput{KwOutput}{Output}  
	\SetKwInput{KwLoss}{Loss}  
	\DontPrintSemicolon
	\LinesNotNumbered
	\KwInput{
		$\D^\text{train}_\text{in}$  \tcp*{Train Set} 
		${f_{\theta}^+}$, $m$  \tcp*{Model, Number of epochs}
		$\mathcal{O}$ = \{contrast, $\cdots$, frost\} \tcp*{Corruption set}
		$\mathcal{S}$ = \{1, 2, 3, 4, 5\} \tcp*{Severity}
		$K+1$ \tcp*{Number of Classes}
	}
	\KwOutput{Trained Model $F^+$}
	\SetKwFunction{FSum}{CnCDataGen} 
	
	\SetKwProg{Fn}{Function}{}{}
	\Fn{\FSum{X$_{ID}$}}{
		X$_\text{pbcc} \leftarrow$ Apply PBCC on X$_{ID}$ as described in Sec. 3.2 of main manuscript\;
		X$^\text{corr}_\text{pbcc} \leftarrow$ $\emptyset$ \;
		\ForEach{x in X$_\text{pbcc}$}{
			\textbf{op} $\sim \mathcal{O}$ \tcp*{Sample Corruptions}  
			\textbf{sev} $\sim \mathcal{S}$ \tcp*{Sample Severity}
			$x_\text{corr} \leftarrow \psi$($x$, \textbf{op}, \textbf{sev})~\cite{hendrycks2018benchmarking} \tcp{Augments $x$ with corruption \textbf{op} of severity \textbf{sev}}
			$X^\text{corr}_\text{pbcc} \leftarrow$ \{X$^\text{corr}_\text{pbcc} \cup x_\text{corr}$\}\;
		}\;
		\KwRet X$^{corr}_{pbcc}$ 
	}
	
	\For{t = 1, 2, $\cdots$, m}{ \tcp{Training Loop}
		\ForEach{ $data$  in $\D^\text{train}_\text{in}$}{          \tcc{$data =$ (image, label)}
			X$_{ID}, \Y$ = $data$ \tcp*{X$_{ID}$ represents a batch of ID images} \;
			X$^\text{corr}_\text{pbcc}$ = \FSum{X$_{ID}$}\;
			$\mathbf{\mathcal{L}}$ = $\lce$(X$_{ID}, \Y$;$f_{\theta}^+$) + $\lce$(X$^\text{corr}_\text{pbcc}$, $K$+1;$f_{\theta}^+$)\;  
			Update model parameters $\theta$
		}
		
	}
	\caption{Compounded Corruptions (CnC)}
	\label{alg:cnc}
\end{algorithm}


\section{Description of State-of-the-art Methods}
\label{sec:ood_detect}

Due to the restrictions on the length, we could not include a description of the state-of-the-art techniques compared with in the main manuscript. However, to keep the manuscript self-contained we include a brief description of each of the compared method here.

\paragraph{Maximum Softmax Probability (MSP) \cite{hendrycks2016baseline}} uses the maximum value corresponding to softmax probability vector, $\max_{i}F_{i}(x)$ as confidence scores to detect OOD examples, where $F(x)$ is the softmax output of the neural network.

\paragraph{ODIN \cite{liang2018enhancing}} ODIN utilizes the calibrated confidence scores computed using temperature scaling and input perturbations for \ood detection. For this study, we set temperature parameter, T = 1000 for all experiments, since the gain gets saturated after T = 1000~\cite{liang2018enhancing}. Perturbation Magnitude $\eta$ is chosen by validating on 1000 images randomly sampled from in-distribution test set ($\mathcal{D}^{test}_{in}$). While using DenseNet as feature extractor, we set $\eta$ = 0.0016 when ID is CIFAR-10, $\eta$ = 0.0012 when ID is CIFAR-100 and $\eta$ = 0.0006 when ID is SVHN. For ResNet, we set $\eta$ = 0.0014 for CIFAR-10, we set $\eta$ = 0.0028 for CIFAR-100 and we set $\eta$ = 0.0014 for SVHN.

\paragraph{Mahalanobis \cite{lee2018simple}} uses Mahalanobis distance-based confidence scores for \ood detection. For fine-tuning the perturbation magnitude $\eta$, we train a Logistic Regression model using 1000 randomly sampled examples from ID test set ($\mathcal{D}^{test}_{in}$) and adversarial examples generated by applying FGSM~\cite{goodfellow2014explaining} on them with a perturbation of size 0.05. $\eta$ is chosen from \{0.0, 0.01, 0.005, 0.002, 0.0014, 0.001, 0.0005\} such that it optimizes the TNR@TPR95.

\paragraph{Generalized-ODIN \cite{hsu2020generalized}} Hsu \etal propose a decomposed confidence model for the purpose of \ood detection, where the logits($f$) of a classifier is  defined using a dividend/divisor structure as follows: $	f(x) =\frac{h(x)}{g(x)}$, where $h(x)$ and $g(x)$ are two functions. The authors propose three variants of \ood detectors, namely, DeConf-I, DeConf-E and DeConf-C which uses Inner-Product, Negative Euclidean Distance and Cosine Similarity for defining $h(x)$ respectively. In this study, we use the DeConf-C variant, since it is shown to be the most robust of all the variants. For calculating \ood score we use the output of $h(x)$ function.
\paragraph{Gram Matrices \cite{sastry2020detecting}} use the concept of Gram-Matrices to compute class-conditional bounds of feature correlations at multiple layers of the network. Given a test image, they compute the total deviation in terms of class conditional layer-wise deviations ($\delta_{l}$) as: $\delta = \sum_{l} \alpha_{l}\delta_{l}$

The value of $\alpha_{l}$ is computed using a validation set of training data.
\paragraph{MixUp \cite{zhang2017mixup}} MixUp works on the principle of vicinical risk minimisation to enhance the support of training distribution through a linear interpolation in the input space with similar interpolation in the associated target space. This improves model robustness to corrupt labels, avoids overfitting. 
\section{Training Details}
\label{sec:trainDet}
\textbf{Implementation Details of CnC:} It is important to note that PBCC and corruption operations are performed on the normalized image tensor.
We train the DenseNet model for 300 epochs with batch size of 64 and weight decay of 0.0001, whereas ResNet-34 is trained with batch size 128 for 200 epochs with weight decay of 0.0005. We use SGD with momentum of 0.9, and initial learning rate starts with $0.1$ and decreases by a factor of 0.1 at 50\% and 75\% of the training epochs.
\paragraph{Description of Evaluation Metrics} 
We denote a sample as positive if it is \id, and negative for \ood samples. Accordingly, we use  TP, TN, FP, FN to denote true positives, true negatives, false positives, and false negatives, respectively. The `R' suffix in any of the above refers to the `rate' of the quantity. We also report test accuracy on \id classes. We compare the performance of various approaches using the following metrics: 
\begin{enumerate}
	\item \textbf{TNR@TPR95:} shows the TNR at 95\% TPR, where TPR = TP / (TP+FN), and TNR = TN /(FP+TN). The \ood detectors aim to attain a high TNR@TPR95, i.e., fraction of \ood samples correctly classified when 95\% of \id samples are correctly classified as \id.
	\item \textbf{AUROC:} is the Area Under the Receiver Operating Characteristic curve and computed as the area under the FPR against TPR curve.
	\item \textbf{Detection Error:} measures the minimum mis-classification probability under the assumption that positive and negative examples have equal probability in the test set. 
\end{enumerate}

\begin{figure}[t]
	\begin{center}
		\includegraphics[width= 0.8\linewidth]{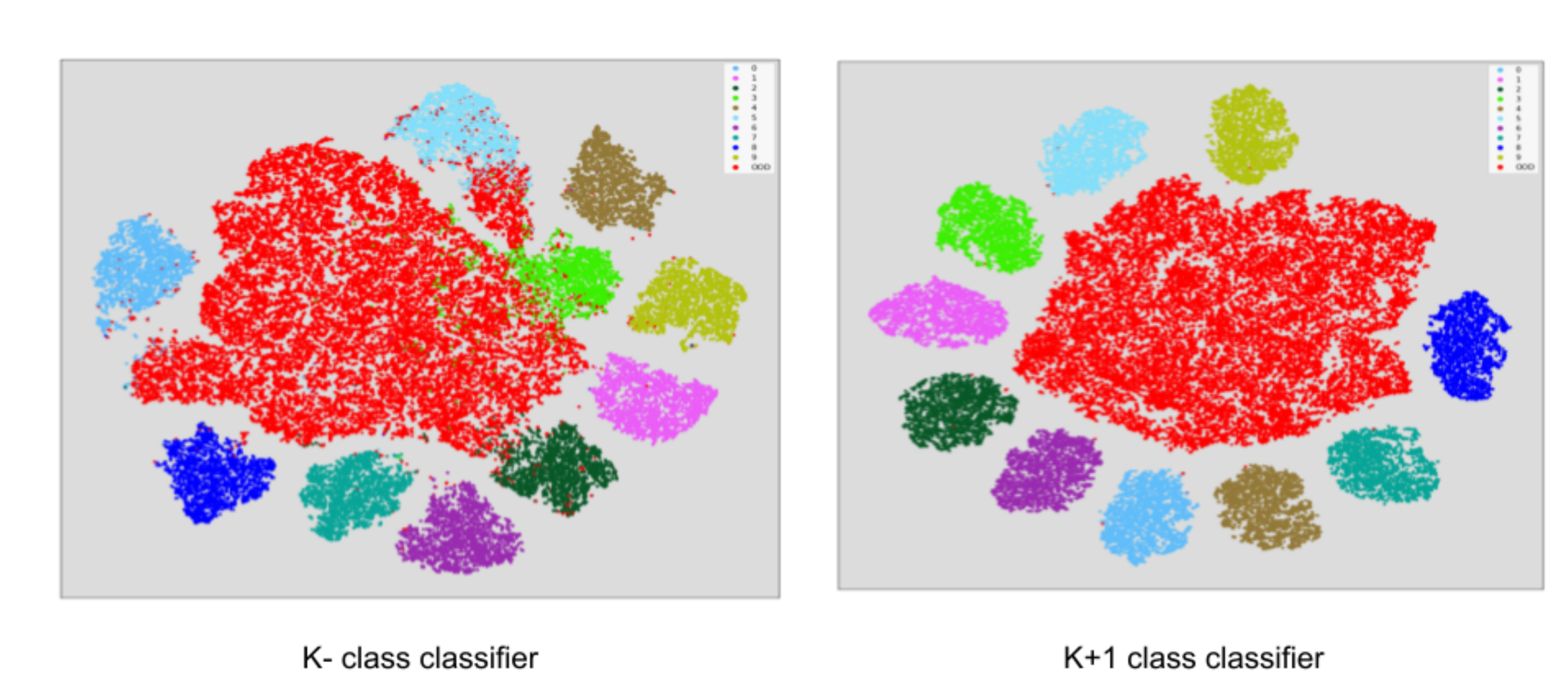}
		\caption{To understand the choice of $(K+1)$ over a $K$ class classifier for OOD detection: we show sample t-SNE plots for both types of classifiers, where CIFAR-10 is used as ID and SVHN is used as \ood(marked in red). The K-class classifier uses temperature scaling \cite{guo2017calibration}, where the temperature parameter is tuned on SVHN test set. On the other hand, the $(K+1)$ class classifier uses SVHN data for $(K+1)^\text{th}$ class during training. The visualization shows that the \ood data is better separated in a (K+1)-class classifier as compared to a K-class classifier. We used ResNet-34 as feature extractor for the experiment.}
		\label{fig:K+1vsK}
	\end{center}
\end{figure}

\begin{figure}[t]
	\begin{center}
		\includegraphics[width=0.7\textwidth, height=0.4\textwidth]{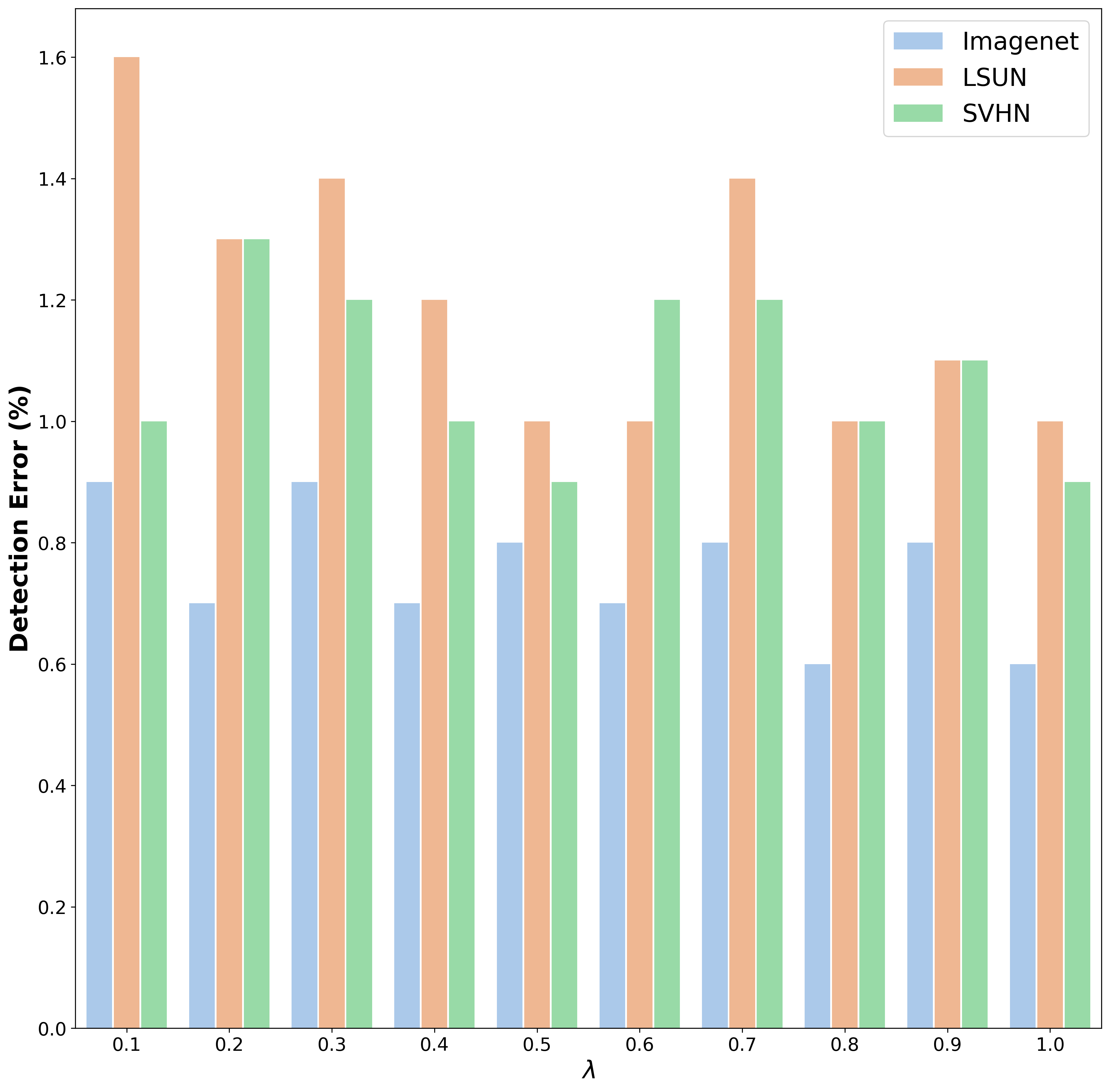}
		\caption{Ablation on $\alpha$ (Eq. 2 in the main paper): Detection Error of the proposed method, CnC, when $\alpha$ is varied in optimisation objective such that it minimises the OOD detection error (main manuscript): $\alpha$ is varied from $0.1$ to $1$. We use DenseNet-BC network and CIFAR-10 dataset as ID for the experiment. We obtain the lowest value of detection error for $\alpha = 1$, when tested with Imagenet, LSUN and SVHN as the \ood datasets.}
		\label{fig:alpha}
	\end{center}
\end{figure}

\section{Comparison with other data synthesis configurations}
Table~\ref{tab:ablation_diff_config} compares performance of models trained using CnC data against the ones trained
%
\begin{table}
\small
\centering
\begin{tabular}{clccc}
\toprule
$\mathcal{D}_\text{out}^\text{test}$ & \textbf{Method} & \textbf{TNR@95TPR} $\pmb{\uparrow}$ & \textbf{AUROC} $\pmb{\uparrow}$ & \textbf{DetErr} $\pmb{\downarrow}$ 
\\ 
\\ 
\midrule
\multirow{3}{*}{$\mathbf{TINc}$} & PBCC  & 93.7 & 98.7 & 5.4  \\
& Corruption  & 96.7 & 99.2 & 4.0  \\
& r-CnC  & 94.8 & 98.7 & 4.5  \\
& CnC & \textbf{99.7} & \textbf{99.9} & \textbf{1.3}  \\
\midrule
\multirow{3}{*}{$\mathbf{LSUNc}$} & PBCC  & 67.2 & 90.3 & 18.6  \\
& Corruption  & 96.7 & 99.2 & 4.1  \\
& r-CnC  & 95.4 & 98.7 & 5.9  \\
& CnC & \textbf{98.3} & \textbf{99.6} & \textbf{3.1}  \\
\bottomrule
\end{tabular}%
\caption{\small Comparison of various possible configurations of our technique using a DenseNet architecture trained on CIFAR-100 as \id dataset. Recall, r-CnC denotes applying corruptions followed by PBCC. \textbf{See suppl. for ablation study.}}
\label{tab:ablation_diff_config}
\end{table}

Using only PBCC, only Corruptions, and r-CnC generated data. Note that in r-CnC we reverse the order of PBCC and corruptions operations. In supplementary, we also compare the UMAP visualization of \ood samples generated using CnC and r-CnC.  Intuitively, the embedding vectors after applying only corruptions remain close to the clusters of \id samples. Hence applying PBCC on top of the corrupted point simply generates \ood samples in the line connecting the two points that does not span a wider area outside \id clusters, and are not much useful for training an \ood detector. In contrast, CnC synthesized \ood data exhibits higher diversity in the embedding space making CnC desirable. 

\section{Using a K vs. a (K+1) Class Classifier}
\label{sec:kplus1VsK}

\paragraph{Why K+1 class classifier?} DNNs are easily fooled with images which humans do not consider meaningful because the closed set nature forces them to choose from one of the known $K$ classes leading to such artifacts which prove to be catastrophic. Recognition in the real world is open set meaning the recognition system should reject novel/unknown/unseen/reject classes at test time. Related work have shown how to produce “fooling” ~\cite{nguyen2015deep} or “rubbish”~\cite{goodfellow2014explaining} images. Even when the images are visually far from the desired class, they produce high confidence scores \cite{hein2019relu}. \textbf {This suggests that thresholding on uncertainty is not sufficient to determine what is unknown/novel class  ~\cite{goodfellow2014explaining} .} Thus, we propose CnC utilizing $(K+1)$ class training with an \ood class.

\paragraph{How do we differ from prior approaches using K+1 classifier?} our approach is inspired by concepts from \cite{Scheirer_2014_TPAMIb}, however, distinctively different from theirs;  \cite{bendale2015open} propose a new layer, Openmax, but ours is a data augmentation scheme. ATOM\cite{chen2021atom} tries to overcome the brittleness of DNN models in the open-world by proposing a (K+1) way classifier. However, they use auxiliary data for effective outlier mining. They argue that using all examples in an auxiliary OOD data for training a classifier can be misleading because of the presence of samples which are non-informative in nature. Recently~\cite{Mohseni_Pitale_Yadawa_Wang_2020} propose to extend the last layer of a deep neural net with K additional nodes specifically for \ood detection. However, unlike our proposed method their training is based on using additional validation set for \ood data along with \id data. We hypothesize that training a K-way classifier using auxiliary data generated using CnC method can hurt  the decision boundary of the classifier and we empirically show the same.

\paragraph{Visualisation of $K$ Vs. $K+1$ class classifier} We show that introducing a $K+1^{th}$ class for OOD detection instead of K-way classifier design does not increase in the number of false negatives (ID classified as OOD). For example, when ID is CIFAR-100 and \ood is SVHN, using DenseNet architecture, a confusion matrix is shown in Table \ref{tab:confusion_matrix_cnc} and Table \ref{tab:confusion_matrix_msp}. Table~\ref{tab:confusion_matrix_msp} shows the confusion matrix for Maximum Softmax Probability (MSP)~\cite{hendrycks2016baseline} \ood detector that uses a K-way classifier. On other hand, Table~\ref{tab:confusion_matrix_cnc} shows the confusion matrix for CnC trained \ood detector which uses a $K+1$ way classifier. We can clearly observe that there is a decrease in number of False Positives when a CnC trained K+1 way model is used, as compared to K-way classifier. 
Refer Figure \ref{fig:K+1vsK} to understand how the \ood data is better separated in a (K+1)-class classifier as compared to a K-class classifier. 

\begin{figure}
	\centering
	\includegraphics[width=\linewidth, height =0.5\linewidth]{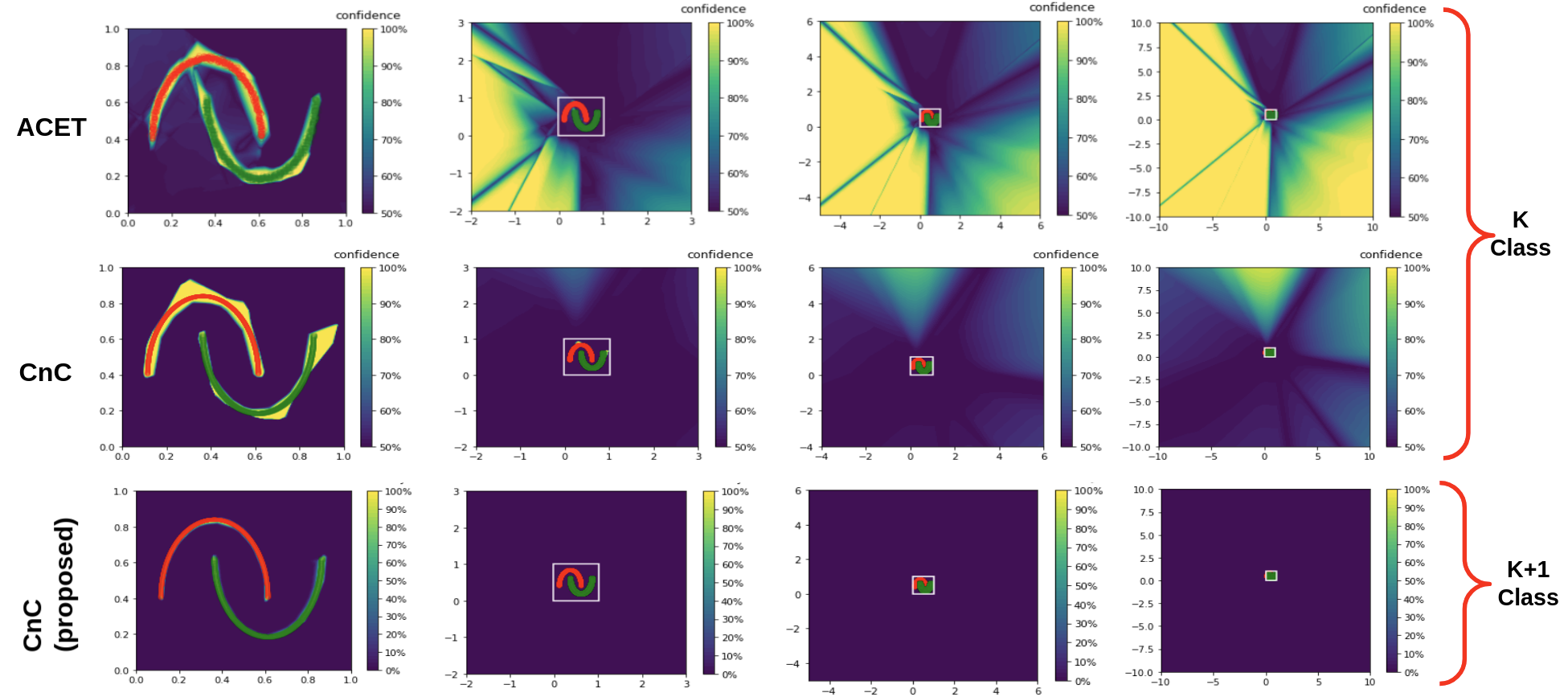}
	\caption{Confidence plot of a NN trained using CnC data on data points far away from \id data. \textbf{Yellow} indicates high confidence, and \textbf{Purple} implies low confidence. \textbf{Row 1:} is output of ACET (K class classifier) \cite{hein2019relu}, \textbf{Row 2:} is for the same $K$-class classifier, but using CnC data during training, whereas \textbf{Row 3:}  is for CnC trained $(K+1)$ class classifier. We observe that the confidence plots for CnC are in general better; those for $(K+1)$ class classifier are better than $K$-class one.} 
	\label{fig:uncertainty}
\end{figure}

ACET \cite{hein2019relu} report that  despite adversarial training confidence of a ReLU based network far away from the \id points remains high. For a given input ${\mathbf{x}}$, we obtain ${\mathbf{y}}$, where ${\mathbf{y}}$ is the Softmax output from a trained NN model $f_{\theta}(\textbf{x})$. We define confidence measure as ${\max(\textbf{y}[:2])}$ (max of \id softmax values). First row of the Fig. \ref{fig:uncertainty} shows the confidence plot for the half moon dataset, with the training data for positive(Red) and negative classes (Green). Yellow color shows high confidence, and purple show the low confidence. Note that the scale of the inputs changes from left to right. As one can see for ACET the classifier indeed outputs low confidence near the \id region but as the scale increases (far from the \id points), we see more and more high confidence regions, showing the fragility of the \ood detection based on the confidence scores. In the second row we show the $K$ class classifier trained with CnC generated data. As we can that the confidence is much lower even in the far off points. We see that the classifier confidence further improves (low in far away regions) with $(K+1)$ class classifier and CnC based training. Note that for $(K+1)$ class classification, though the classification is $K$-way, the confidence depicted is still ${\max(\textbf{y}[:K])}$. The Fig. \ref{fig:uncertainty} serves to illustrate (1) the benefit of CnC for $K$ class classification, and (2) merit of using $(K+1)$ over a $K$ class classifier.

\begin{table}[H]
\centering 
    
    \begin{tabular}{cccc}
         &  & \multicolumn{2}{c}{$\mathbf{Actual}$} \\
         \cline{3-4}
         & & ID & OOD \\
         \cline{3-4}
         \multirow{4}{*}{\begin{turn}{90}$\mathbf{Predicted}$\end{turn}} & \begin{turn}{0}ID\end{turn} & True Positive = 9514 & False Positive = 151\\
          & & (TP) & (FP) \\
          \cline{3-4}
         & \begin{turn}{0}OOD\end{turn} & False Negative = 486 & True Negative = 9849 \\
          & & (FN) & (TN) \\
         \cline{3-4}\\
    
    \end{tabular}
    
    \caption{Confusion Matrix for K+1 way CnC trained Classifier}
    \label{tab:confusion_matrix_cnc}
\end{table}

\begin{table}[H]
    \centering\centering   
    \begin{tabular}{cccc}
         &  & \multicolumn{2}{c}{$\mathbf{Actual}$} \\
         \cline{3-4}
         & & ID & OOD \\
         \cline{3-4}
         \multirow{4}{*}{\begin{turn}{90}$\mathbf{Predicted}$\end{turn}} & \begin{turn}{0}ID\end{turn} & True Positive = 9354 & False Positive = 7732\\
          & & (TP) & (FP) \\
          \cline{3-4}
         & \begin{turn}{0}OOD\end{turn} & False Negative = 646 & True Negative = 2268 \\
          & & (FN) & (TN) \\
         \cline{3-4}\\
    \end{tabular}
    
    \caption{Confusion Matrix for K-way MSP~\cite{hendrycks2016baseline} trained Classifier}
    \label{tab:confusion_matrix_msp}
\end{table}


\begin{figure*}
\centering
\begin{subfigure}{\linewidth}
	\centering
	\includegraphics[width=0.7\linewidth]{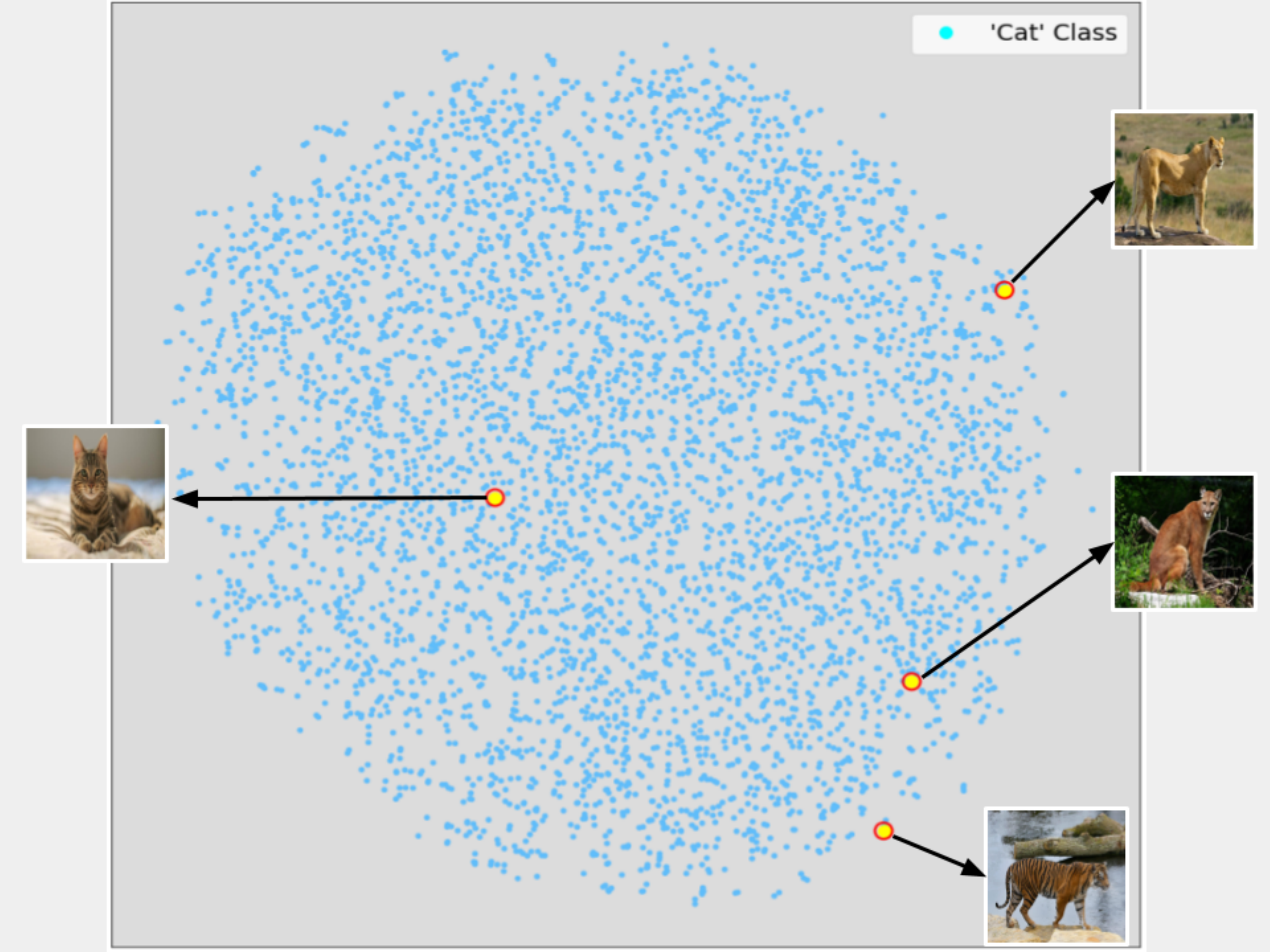}  
\end{subfigure}
\begin{subfigure}{\linewidth}
	\centering
	\includegraphics[width=0.7\linewidth]{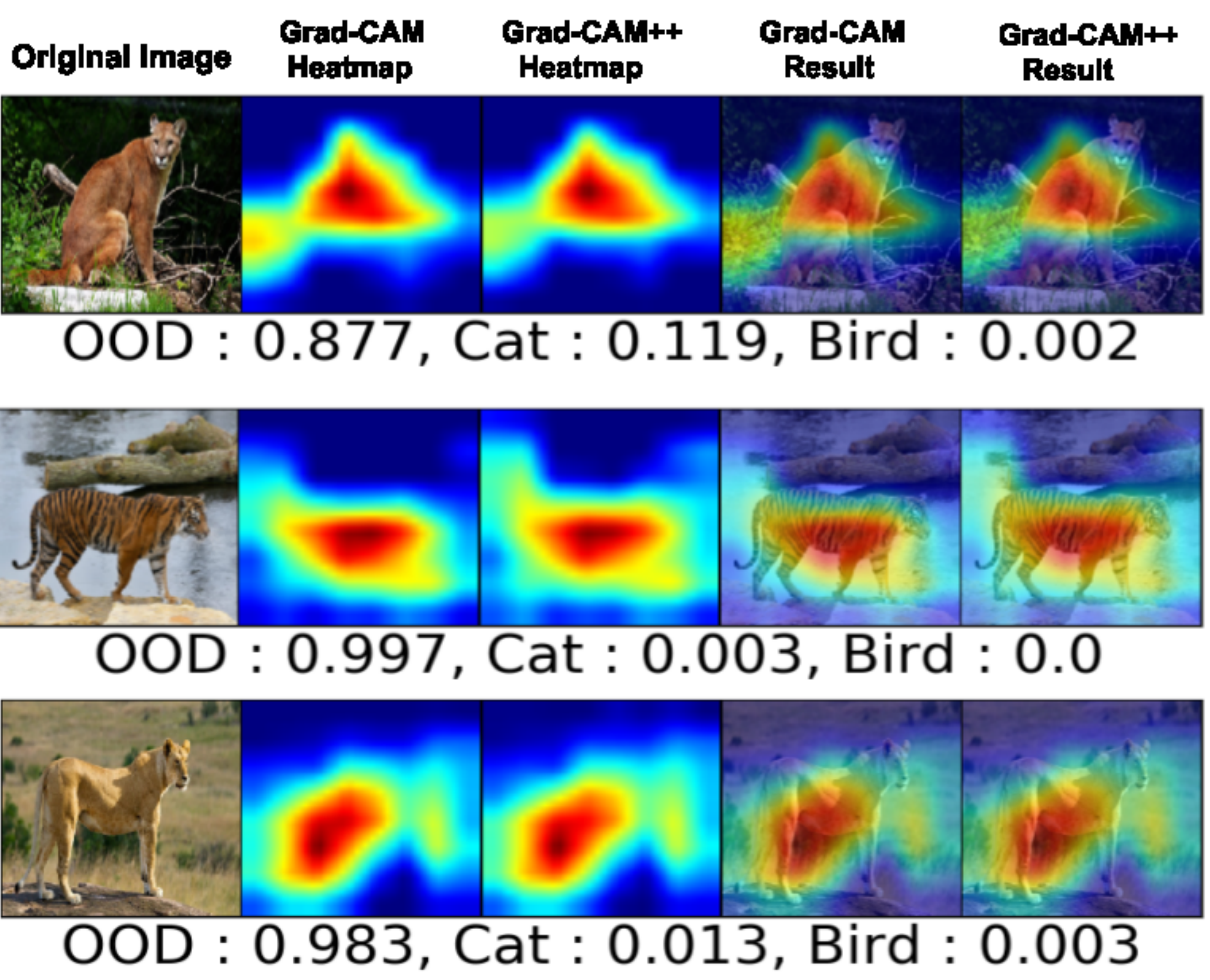}  
\end{subfigure}
\caption{We study the efficacy of CnC trained model on boundary cases. (a) t-SNE mapping of images that are similar to "Cat" class of CIFAR-10 but are semantically a different class: Examples shown are cougar, tiger and lioness.We test the CnC model trained on Cat class on a semantically similar but a different class such as \texttt{cougar}, \texttt{tiger} and \texttt{lioness} downloaded from internet. (b) GradCAM~\cite{gradcam} and GradCAM++~\cite{gradcam++} visualizations of corresponding images. Note that CnC is able to successfully classify them as OOD with high probability $0.877$, $0.997$, and $0.983$ }
\label{fig:cougar}
\end{figure*}

\renewcommand{\arraystretch}{1.5}
\begin{table}
\setlength{\tabcolsep}{0.2\tabcolsep}
\centering 
\begin{tabular}{c|l|ccc}
\toprule
\multirow{2}{*}{$\mathcal{D}_\text{in}^\text{train}$} & 
\multirow{2}{*}{Range ($\lambda$)} & \textbf{TNR@TPR95} & \textbf{AUROC} & \textbf{DetErr} 
\\ 
&  & $\pmb{\uparrow}$ & $\pmb{\uparrow}$ & $\pmb{\downarrow}$  
\\ 
\toprule
\multirow{5}{*}{\begin{turn}{90}$\mathbf{CIFAR-10}$\end{turn}} 
& \textbf{U(0.0, 1.0)} & \textbf{99.8} & \textbf{99.9} & \textbf{0.9}  \\
& $U(0.1,0.9)$  & 84.1 & 96.5 & 9.0  \\
& $U(0.2,0.8)$ & 79.3 & 96.5 & 9.5  \\
& $U(0.3,0.7)$ & 78.9 & 96.4 & 9.8  \\
& $U(0.4,0.6)$ & 77.8 & 96.2 & 10.3  \\
\toprule
\end{tabular}

\caption{Ablation of sampling range [$\lambda, (1-\lambda$)] on uniform distribution (U) to ensure maximum number of CnC generated OOD samples do not lie on ID regions. Recall from the main manuscript, the cut-out region is determined by $(1-\lambda)$ for bounding box in our proposed algorithm. We use DenseNet as network architecture, CIFAR-10 as ID dataset and Tiny-ImageNet(Crop), LSUN(Crop) and SVHN as OOD Dataset for all methods.} 
\label{tab_uniform_dist_lambda}
\end{table}



\begin{table}
\centering
\begin{tabular}{c|l|c|c|c}
\hline
& \multicolumn{2}{c|}{RMS CE $\downarrow$} & \multicolumn{2}{c}{ECE CE $\downarrow$}\\
\hline
$\mathcal{D}_{in}$ & TS & CnC & TS & CnC \\
\cline{2-5}
CIFAR-10 & 11.09 & \textbf{6.18} & 5.12 & \textbf{3.15}\\

CIFAR-100 & 26.01 & \textbf{19.36} & 19.57 & \textbf{15.4}\\
\hline
\end{tabular}
\captionof{table}{Calibration Results for temperature tuned baseline and CnC trained model. Lower calibration error is better.}
\label{tab:1}
\end{table}

In Table \ref{tab:1}, we compare TS against our proposed CnC trained model in terms of calibration. For TS method, we find optimal temperature by tuning on validation set of ID datasets and for CnC we keep T=1 (original). The proposed method, CnC achieves better calibration than TS, thus making it more reliable.

Additionally, our proposed method is better than TS in terms of robustness as we train on several types of corruptions (proxy for \ood). 
Note: ODIN~\cite{liang2018enhancing} uses temperature scaling for \ood detection, however, \ood test sets have significantly different distributions than the training distribution. Hence in this case TS is not guaranteed to produce a calibrated probability. Guo \etal explicitly mention temperature scaling fails in the case of adversarial setting or when test samples are noisy~\cite{guo2017calibration}.

\section{Effect of Diversity on OOD Detection}
\label{sec:div-entropy}

We adapt data diversity from Zhang et al\cite{zhang2019active} to measure OOD diversity. Refer to Algorithm \ref{alg:OODDiv} to compute diversity. 
For computing entropy, we use $H(x)  = -\sum_{n=1}^{K} p_i \log(p_i)$. Here $p_i$ is the model-generated probability that the data point belongs to class $i$, and $K$ is he number of classes. As reported in the main manuscript, we observe that higher entropy and higher diversity leads to better OOD detection.





The objective of this experiment is to corroborate the fact that higher the diversity of generated \ood samples, better the \ood detection performance measured using metrics such as AUROC, TNR@TPR95 and Detection Error. To this end, we train two (K+1) class classifiers, in first we generate proxy \ood samples by applying \emph{Gaussian} noise and similarly in second  we use \emph{Speckle} noise as corruptions. The parameters of Gaussian/Speckle noise are same as the default settings proposed in Hendrycks and Dietterich \url{https://github.com/hendrycks/robustness} 

We compute the mean diversity of proxy \ood samples generated using Speckle noise and Gaussian noise. We find the diversity of Speckle noise as $2.16$ and that of the Gaussian counter part to be $1.71$ using Algorithm  \ref{alg:OODDiv}. In Table~\ref{tab:densenetgaussianvspeckle}, we show that Speckle noise trained \ood detector performs better as compared to Gaussian trained \ood detector, thus, providing an evidence to our hypothesis, higher diversity of OOD points yields better OOD detection.




\begin{table}
 \centering
 \tiny

 \begin{tabular}{l|ll|ccc}
\toprule
 \multirow{2}{*}{$\mathcal{D}_\text{in}^\text{train}$} & \multirow{2}{*}{$\mathcal{D}_\text{out}^\text{test}$} & 
 \multirow{2}{*}{\textbf{Method}} & \textbf{TNR (95\% TPR)} & \textbf{AUROC} & \textbf{DetErr} 
 \\ 
  & & & $\pmb{\uparrow}$ & $\pmb{\uparrow}$ & $\pmb{\downarrow}$  \\ 
 \toprule
 \multirow{12}{*}{\begin{turn}{90}$\mathbf{CIFAR-10}$\end{turn}} & \multirow{2}{*}{$\mathbf{TINc}$} & Gaussian Noise & 75.8 & 93.6 & 12.1  \\
 & & Speckle Noise & \textbf{95.2} & \textbf{98.8} & \textbf{4.7} \\ 
 \cline{2-6}
& \multirow{2}{*}{$\mathbf{TINr}$} & Gaussian Noise & 72.5 & 93.1 & 13  \\
& & Speckle Noise & \textbf{97.5} & \textbf{99.4} & \textbf{2.8}  \\
 \cline{2-6}

& \multirow{2}{*}{$\mathbf{LSUNc}$} & Gaussian Noise & 77 & 92.2 & 12.3  \\
& & Speckle Noise & \textbf{80.2} & \textbf{94.9} & \textbf{10.6}     \\
 \cline{2-6}

& \multirow{2}{*}{$\mathbf{LSUNr}$} & Gaussian Noise & 83.1 & 96.1 & 9.5  \\
& & Speckle Noise & \textbf{98.5} & \textbf{99.6} & \textbf{2.1}  \\
 \cline{2-6}

& \multirow{2}{*}{$\mathbf{iSUN}$} & Gaussian Noise & 77.4 & 94.5 & 11.3  \\
& & Speckle Noise & \textbf{97.1} & \textbf{99.4} & \textbf{3.2}  \\
 \cline{2-6}

& \multirow{2}{*}{$\mathbf{SVHN}$} & Gaussian Noise & 78.6 & 93.1 & 13  \\
& & Speckle Noise & \textbf{97.5} & \textbf{99.4} & \textbf{2.8}  \\
\bottomrule
\end{tabular}
\caption{Comparison between Gaussian and Speckle noise trained \ood classifiers. We use DenseNet as network architecture for all methods.}
\label{tab:densenetgaussianvspeckle}
\end{table}

\noindent\setlength{\tabcolsep}{0.4\tabcolsep}
\section{Subfunction Error bound}
\label{sec:suplSuberror}
In this section, we briefly explain the empirical error bounds for subfunction as provided in \cite{xuji} for the sake of completeness/self sufficiency. Consider a distribution $D_S$ over size-N datasets that have the same activation region data distributions and dataset sizes as $S$, where $S$ is the training dataset $(\mathcal{X} \times \mathcal{Y})^N$ drawn \iid from $D_S$. Then for all $i \in [1,C]$, for any given $\delta \in (0,1]$, with probability > $1-\delta$ :
\begin{equation*}
    \underbrace{R^{*}_{D_S} (h_i) \triangleq \mathbb{E}_S[\hat{R}^{*}_{S} (h_i)]}_{\text{expected smooth error}} \leq
    \overbrace{\underbrace{\hat{R}^{*}_{S} (h_i)}_{\text{empirical smooth error}} + \underbrace{\cfrac{1}{\mathbb{P}(h_i)}\sqrt{\cfrac{log\frac{2}{\delta}}{2N}}}_{\text{generalization gap}}}^{\text{bound on expected smooth error}}
\end{equation*}

where,
\begin{equation*}
    \mathbb{P}(h_i) \triangleq \cfrac{\sum_{j\in[1,C]}N_j k(h_i,h_j)}{\sum_{l\in[1,C]}\sum_{j\in[1,C]}N_j k(h_l,h_j)}
\end{equation*}
for any non-negative weighting distance function between subfunctions, $k : H \times H \rightarrow \mathbb{R}^{\geq 0}$. \textbf{Note:} C varies as $1/sqrt(N)$ where C is number of polytopes and N the number of training samples.

\section{Time Efficiency}
For  applications  demanding  real-time  performance, it is crucial to have low latency in systems using \dnn for inference. To analyze the comparative efficiency, we run the experiments on a desktop system with Intel Core i7 (9th Gen) processor, $16$GB RAM, and NVidia RTX 2060 GPU . We report inference time in \emph{ms} averaged over $10K$ images at test time. CnC took the least time when compared to other \sota techniques. We report that both MSP \cite{hendrycks2016baseline} and CnC model took $0.7$ms whereas ODIN \cite{liang2018enhancing} and Generalised-ODIN \cite{hsu2020generalized} took $3.1$ms and $3.3$ms respectively. Inference time of Mahalanobis~\cite{lee2018simple} was $8.8$ms while GRAM matrices\cite{sastry2020detecting} took $10$ms. We use CIFAR10 trained models with \cite{huang2018densely} as the feature extractor for this experiment.

\section{Summary performance of various approaches}

As is a practice in the literature we show the summary performance of various approaches, along with their key conceptual differences as shown in Tab. \ref{tab:summary_results_conceptual_comparison}

\begin{table}
	\centering
	\setlength{\tabcolsep}{5pt}
		\begin{tabular}{lcccc}\\
			\toprule[1pt]
			Method & IP & OV & RT \\
			\midrule[0.65pt]
			MSP (ICLR'17)  \cite{hendrycks2016baseline} & $\times$ & $\times$ & $\times$ \\
			ODIN (ICLR'18) \cite{liang2018enhancing} & $\surd$ & $\surd$ & $\times$ \\
			Gen-ODIN (CVPR'21) \cite{hsu2020generalized} & $\surd$ & $\times$ & $\surd$ \\
			Maha (NeurIPS'18) \cite{lee2018simple} & $\times$ & $\surd$ & $\times$ \\
			Gram Matrices (ICML'20)\cite{sastry2020detecting} & $\times$ & $\times$ & $\times$\\
			ACET (CVPR'20) \cite{hein2019relu}  & $\times$ & $\times$ & $\surd$  \\
			ATOM (ECML'21) \cite{chen2021atom} &  $\times$ & $\times$  & $\surd$   \\
			CnC(Proposed) & $\times$ & $\times$ & $\surd$ \\
			\bottomrule[1pt]
		\end{tabular}
		\caption{ We show the conceptual differences between various techniques. Notation: \textbf{IP} indicates whether a method needs input pre-processing or not during inference, \textbf{OV} marks those methods which requires \ood validation set for fine-tuning hyper-parameters, and \textbf{RT} defines whether re-training of neural network is required or not. }
\label{tab:summary_results_conceptual_comparison}
\end{table}


\section{Experiments on More Datasets}
\label{sec:suplRob}

We compare our proposed method against existing literature on distribution shift benchmark datasets such as Imagenet-O  and Imagenet-R. We also show empirically that our proposed method outperforms the existing literature by a large margin when tested against corrupted \ood dataset such as Tiny-ImageNet-C.

\paragraph{ImageNet-O} ImageNet-O is a natural adversarial out-of-distribution dataset curated in \cite{hendrycks2019nae}. It consists of images which are
natural, unmodified, real-world examples and are selected to cause a model to make a mistake, as with synthetic adversarial examples. The design is such that it causes models to mistake the examples as high-confidence in-distribution examples. It allows us to estimate model uncertainty due to data distribution shift. It comprises of 2000 test images which occur naturally but mimic synthetic adversarial examples.
\paragraph{ImageNet-R.} ImageNet-R is a $30000$ image test set containing various renditions of 200 ImageNet object classes. These renditions belong to either of following categories - art, cartoons, graffiti, embroidery, graphics, origami, paintings, patterns, plastic objects, plush objects, sculptures, sketches, tattoos, toys, and video games.

\paragraph{Corrupted \ood} To mimic \ood occurring more naturally in physical world, we also compare against TinyImageNet-C , which consists of corrupted versions of images in Tiny-ImageNet. For fair comparison, in this study, we use CnC-3 for testing instead of CnC which uses all types of corruptions in training. Also, during testing we remove those corruption types from TinyImagenet-C which were used for training in CnC-3, specifically \texttt{contrast}, \texttt{frost} and \texttt{saturate} which are determined empirically to show higher diversity in embedding space than individual noise types considered in isolation. We strictly maintain that training and testing corruption sets are exhaustive and mutually exclusive.


While testing, images in ImageNet-O, ImageNet-R and TinyImageNet-C are down-sampled to a size of $32 \times 32$. Table ~\ref{tab:robustness1}, Table 4 (main), and Table~\ref{tab:robustness3} present experimental results that back CnC for robustness. We note that CnC not only excels against natural \ood transforms but also is resistant to certain adversarial perturbations and corruption transforms. 
\begin{table}
\centering
\centering
\setlength{\tabcolsep}{0.7\tabcolsep}
 \begin{tabular}{l|ccc}
\toprule
 $\mathbf{Method}$ & $\mathbf{TNR}$ & $\mathbf{AUROC}$ & $\mathbf{DetErr}$\\ 
 
    & $\mathbf{(95\% TPR)}$ & & \\
    & $\pmb{\uparrow}$ & $\pmb{\uparrow}$ & $\pmb{\downarrow}$  \\ 
\toprule
 
   MSP \cite{hendrycks2016baseline}  & 16.3 & 75.1 & 30.3  \\
    ODIN \cite{liang2018enhancing}  & 26.6 & 80.2 & 26.5 \\
   Gen-ODIN \cite{hsu2020generalized} & 43.8 & 86.2 & 21.5  \\
   Mahalanobis \cite{lee2018simple} & 25.9 & 67.9 & 34.5  \\
   Gram Matrices \cite{sastry2020detecting} & 31.28  & 79.86 & 27.4 \\
   \textbf{CnC} & \textbf{44.2} & \textbf{86.9} & \textbf{20.9}\\
 
\bottomrule
\end{tabular}
\caption{Experiment results for $\mathbf{ImageNet-O}$ dataset. We use DenseNet as feature extractor and CIFAR-100 as in-distribution. }
\label{tab:robustness1}
\end{table}

\begin{table}[t]
\centering
\setlength{\tabcolsep}{0.7\tabcolsep}
 \begin{tabular}{l|ccc}
\toprule
 $\mathbf{Method}$ & $\mathbf{TNR}$ & $\mathbf{AUROC}$ & $\mathbf{DetErr}$\\ 
 
    & $\mathbf{(95\% TPR)}$ & & \\
    & $\pmb{\uparrow}$ & $\pmb{\uparrow}$ & $\pmb{\downarrow}$  \\ 
\toprule
 
   MSP \cite{hendrycks2016baseline}  & 20.5 & 77.5 & 28.5  \\
    ODIN \cite{liang2018enhancing}  & 22.9 & 79.4 & 27.3  \\
   Gen-ODIN \cite{hsu2020generalized} & 0.2 & 50.0 & \textbf{18.3}  \\
   Mahalanobis \cite{lee2018simple} & 10.7 & 59.0 & 43.3  \\
   Gram Matrices \cite{sastry2020detecting} & 16.1  & 69.3 & 35.7 \\
   CnC-3 & \textbf{27.8} & \textbf{81.2} & 25.9\\
 
\bottomrule
\end{tabular}
\caption{Experiment results for $\mathbf{Tiny-ImageNet-C}$ dataset. We use DenseNet as feature extractor and CIFAR-100 as in-distribution. The values are averaged over corrupted versions of \ood benchmark datasets. We show that even training with CnC samples generated by applying a subset of corruptions mentioned in \cite{hendrycks2018benchmarking}, we get an almost optimal \ood detection performance.}
\label{tab:robustness3}
\end{table}

\section{Performance on Non-Semantic Data Shift}

In this section, we aim to show the efficacy of CnC algorithm when tested against a non-semantic data shift. For this, we use PACS~\cite{zhou2020deep} dataset, commonly used as benchmark for domain generalization. It consists of four domains, namely Photo (1,670 images), Art Painting (2,048 images), Cartoon (2,344 images) and Sketch (3,929 images), where images are of dimension $227 \times 227$. Each domain contains seven classes which are dog, elephant, giraffe, horse, house, person and guitar. Table~\ref{tab:pacs_cifar10} and Table~\ref{tab:cnc_lim} tabulates the experimental results. Table 4 in the main manuscript presented the performance of \ood detectors under domain shift, extending this, Table \ref{tab:cnc_lim} illustrated that CnC trained on one domain in PACS dataset and tested on the others outperforms Generalised-ODIN \cite{hsu2020generalized}. Similarly results are consistent in Table \ref{tab:pacs_cifar10}.

\begin{table}[t]
\small
 \centering

 \setlength{\tabcolsep}{0.2\tabcolsep}
 
 \begin{tabular}{l|ll|ccc}

\toprule
 $\mathcal{D}_{in}^{train}$ & $\mathcal{D}_{out}^{test}$ & $\mathbf{Method}$ & $\mathbf{TNR}$ & $\mathbf{AUROC}$ & $\mathbf{DetErr}$\\ 
 
  & & & $\mathbf{(95\% TPR)}$ & & \\
  & & & $\pmb{\uparrow}$ & $\pmb{\uparrow}$ & $\pmb{\downarrow}$  \\ 
\toprule
\multirow{12}{*}{\begin{turn}{90}$\mathbf{CIFAR-10}$\end{turn}} & \multirow{6}{*}{$\mathbf{PACS(Dog)}$} & MSP \cite{hendrycks2016baseline} & 37.7 & 87.4 & 18.9  \\
& & ODIN \cite{liang2018enhancing} & 71.1 & 91.7 & 15.0  \\
& & Gen-ODIN \cite{hsu2020generalized} & 74.7 & 93.3 & 14.1  \\
& & Mahalanobis \cite{lee2018simple} & 15.1 & 40.6 & 44.4  \\
& & Gram Matrices  \cite{sastry2020detecting} & 28.6 & 85.1 & 21.5 \\
& & CnC(Proposed) & \textbf{78.5} & \textbf{94.2} & \textbf{11.9}\\
\cline{2-6}
& \multirow{6}{*}{$\mathbf{PACS(Horse)}$} & MSP \cite{hendrycks2016baseline} & 24.1 & 78.3 & 28.3  \\
& & ODIN \cite{liang2018enhancing}  & 52.9  & 87.1 & 20.8  \\
& & Gen-ODIN \cite{hsu2020generalized} & 65.1 & 90.8 & 17.4  \\
& & Mahalanobis \cite{lee2018simple}  & 9.9 & 34.8 & 46.8  \\
& & Gram Matrices \cite{sastry2020detecting} & 46.4  & 88.3 & 18.3 \\
& & CnC(Proposed) & \textbf{73.3} & \textbf{93.7} & \textbf{12.5} \\
\bottomrule
\end{tabular}
\caption{Comparison of CnC with competitive \ood detection methods when a model is trained on CIFAR10 and tested on samples from class \texttt{dog} and \texttt{horse} in PACS\cite{li2017deeper}. To test fairly for non-semantic shift, we only consider the dog and horse class, as CIFAR-10 also has natural images of dog and horse in its training set. We use DenseNet as network architecture for all methods.}
\label{tab:pacs_cifar10}
\end{table} 

\renewcommand{\arraystretch}{1.75}
\begin{table}
\small
\centering
\setlength{\tabcolsep}{0.2\tabcolsep}
\begin{tabular}{c|c|ccc}
\toprule
\multirow{2}{*}{$\mathcal{D}_\text{in}^\text{train}$} & 
\multirow{2}{*}{$\mathcal{D}_\text{out}^\text{test}$} & \textbf{TNR@TPR95} & \textbf{AUROC} & \textbf{DetErr} 
\\ 
& & $\pmb{\uparrow}$ & $\pmb{\uparrow}$ & $\pmb{\downarrow}$  
\\ 
\cline{3-5}
& & \multicolumn{3}{c}{\textbf{CnC} / \textbf{Gen-ODIN}~\cite{hsu2020generalized} }\\
\toprule
\multirow{3}{*}{\begin{turn}{90}$\mathbf{Photo}$\end{turn}} 
& Art and Painting  & \textbf{86.7}/52.7 & \textbf{96.0}/ 88.0 & \textbf{7.7}/19.8  \\
& Sketch  & 15.3/\textbf{99.7} & 57.8/\textbf{99.9} & 38.8/\textbf{1.2}  \\
& Cartoon & 34.1/\textbf{77.2} & 64.5/\textbf{95.0} & 31.3/\textbf{12.0}  \\
\hline
\multirow{3}{*}{\begin{turn}{90}$\mathbf{Cartoon}$\end{turn}} 
& Photo  & \textbf{92.4}/68.2 & \textbf{96.6}/90.2 & \textbf{6.1}/16.8 \\
& Sketch  & \textbf{97.1}/76.2 & \textbf{97.6}/95.4 & \textbf{3.9}/10.9  \\
& Art and Painting & \textbf{98.8}/75.1 & \textbf{99.5}/94.9 & \textbf{2.5}/11.4  \\
\hline
\multirow{3}{*}{\begin{turn}{90}$\mathbf{Sketch}$\end{turn}} 
& Art and Painting  & \textbf{100.0}/99.9 & 100.0/100.0 & \textbf{0.1}/0.2  \\
& Photo & \textbf{100.0}/\textbf{100.0} & \textbf{100.0}/\textbf{100.0} & \textbf{0.1}/\textbf{0.1}  \\
& Cartoon & \textbf{100.0}/99.8 & 100.0/100.0 & \textbf{0.3}/0.5  \\
\hline
\multirow{3}{*}{\begin{turn}{90}\small{$\mathbf{Art \& Painting}$}\end{turn}} 
&  Photo & \textbf{57.3}/30.0 & 83.0/\textbf{89.0} & \textbf{15.4}/17.5  \\
& Sketch  & \textbf{98.9}/17.3 & \textbf{99.6}/76.1 & \textbf{2.8}/29.3  \\
& Cartoon & \textbf{49.1}/31.6 & \textbf{86.5}/72.1 & \textbf{16.1}/18.5  \\
\bottomrule
\end{tabular}
\caption{Results of non-semantic shift, when we train on one domain in PACS dataset~\cite{zhou2020deep} and test on other three. Observe that the CnC trained model performs better than Gen-ODIN~\cite{hsu2020generalized} in certain cases.}
\label{tab:cnc_lim}
\end{table}

\renewcommand{\arraystretch}{1}

\begin{table}
\centering
 \begin{tabular}{l|l|l|ccc}
\toprule
$\mathcal{D}_{in}^{train}$ & $\mathcal{D}_{out}^{test}$ & $\mathbf{Method}$ & $\mathbf{TNR}$ & $\mathbf{AUROC}$ & $\mathbf{DetErr}$\\ 

 & & & $\mathbf{(95\% TPR)}$ & & \\
 & & & $\pmb{\uparrow}$ & $\pmb{\uparrow}$ & $\pmb{\downarrow}$ \\ 
\toprule

\multirow{24}{*}{\begin{turn}{90}$\mathbf{CIFAR-100}$\end{turn}} & \multirow{3}{*}{$\mathbf{TINc}$} & PBCC  & 93.7 & 98.7 & 5.4  \\
& & Corruption  & 96.7 & 99.2 & 4.0  \\
& & r-CnC  & 94.8 & 98.7 & 4.5  \\
& & CnC & \textbf{99.7} & \textbf{99.9} & \textbf{1.3}  \\
 \cline{2-6}
& \multirow{3}{*}{$\mathbf{TINr}$} & PBCC  & 91.0 & 98.3 & 6.9  \\
& & Corruption  & 94.5 & 98.9 & 5.0  \\
& & r-CnC  & 93.2 & 97.9 & 6.1  \\
& & CnC & \textbf{99.6} & \textbf{99.8} & \textbf{1.6}  \\
\cline{2-6}
& \multirow{3}{*}{$\mathbf{LSUNc}$} & PBCC  & 67.2 & 90.3 & 18.6  \\
& & Corruption  & 96.7 & 99.2 & 4.1  \\
& & r-CnC  & 95.4 & 98.7 & 5.9  \\
& & CnC & \textbf{98.3} & \textbf{99.6} & \textbf{3.1}  \\
 \cline{2-6}
& \multirow{3}{*}{$\mathbf{LSUNr}$} & PBCC  & 97.2 & 99.4 & 3.4  \\
& & Corruption  & 95.7 & 99.1 & 4.5  \\
& & r-CnC  & 94.7 & 98.6 & 5.6  \\
& & CnC & \textbf{99.9} & \textbf{99.9} & \textbf{0.9}  \\
 \cline{2-6}
& \multirow{3}{*}{$\mathbf{iSUN}$} & PBCC  & 95.2 & 99 & 4.8  \\
& & Corruption  & 94.0 & 98.8 & 5.3  \\
& & r-CnC  & 93.5 & 97.7 & 4.5  \\
& & CnC & \textbf{99.6} & \textbf{99.9} & \textbf{1.4}  \\
\cline{2-6}
& \multirow{3}{*}{$\mathbf{SVHN}$} & PBCC  & 10.4 & 72.2 & 32.6  \\
& & Corruption  & 98.5 & 99.5 & 2.8  \\
& & r-CnC  & 98.0 & 98.5 & 10.8  \\
& & CnC & \textbf{98.7} & \textbf{99.7} & \textbf{2.0}  \\
 \cline{2-6}
\bottomrule
\end{tabular}
\caption{Ablation: We use DenseNet as network architecture for all methods. CnC denotes our proposed method Corruption Over PBCC and r-CnC denotes PBCCOverCorruption.}
\label{tab:ablation_complete}
\end{table}

\noindent\setlength\tabcolsep{6pt}%
\begin{table}
 \centering
\resizebox{0.9\linewidth}{!}{%
 \begin{tabular}{l|ll|ccc}
\toprule
 $\mathcal{D}_{in}^{train}$ & $\mathcal{D}_{out}^{test}$ & $\mathbf{Method}$ & $\mathbf{TNR}$ & $\mathbf{AUROC}$ & $\mathbf{DetErr}$\\ 
 
  & & & $\mathbf{(95\% TPR)}$ & & \\
  & & & $\pmb{\uparrow}$ & $\pmb{\uparrow}$ & $\pmb{\downarrow}$  \\ 
 \cline{2-6}
 
 \multirow{36}{*}{\begin{turn}{90}$\mathbf{CIFAR-10}$\end{turn}} & \multirow{6}{*}{$\mathbf{TINc}$} & MSP \cite{hendrycks2016baseline}  & 56.7 & 93.8 & 11.9  \\
& & ODIN \cite{liang2018enhancing}   & 96.8 & 99.2 & 4.0  \\
& & Gen-ODIN \cite{hsu2020generalized} & 93.0 & 98.6 & 5.9  \\
& & Mahalanobis \cite{lee2018simple} & 75.0 & 89.6 & 13.7  \\
& & Gram Matrices \cite{sastry2020detecting} & 96.7  & 99.3 & 3.9 \\
& & CnC(Proposed) & \textbf{99.9} & \textbf{100} & \textbf{0.6}\\
 \cline{2-6}
& \multirow{6}{*}{$\mathbf{TINr}$} & MSP  & 58.9 & 94.1 & 11.5  \\
& & ODIN \cite{liang2018enhancing}  & 90.6 & 98.2 & 6.7  \\
& & Gen-ODIN \cite{hsu2020generalized} & 94.9 & 99.0 & 5.0  \\
& & Mahalanobis \cite{lee2018simple} & 84.9 & 94.0 & 9.9  \\
& & Gram Matrices \cite{sastry2020detecting} & 98.8  & 99.7 & 2.1 \\
& & CnC(Proposed) & \textbf{99.9} & \textbf{100.0} & \textbf{0.8} \\
 \cline{2-6}
& \multirow{6}{*}{$\mathbf{LSUNc}$} & MSP  & 51.8 & 92.9 & 13.1  \\
& & ODIN \cite{liang2018enhancing} & 98.1 & 99.5 & 3.3  \\
& & Gen-ODIN \cite{hsu2020generalized} & 92.1 & 98.4 & 6.4  \\
& & Mahalanobis \cite{lee2018simple} & 85.0 & 92.6 & 10.0  \\
& & Gram Matrices \cite{sastry2020detecting} & 88.4  & 97.5 & 8.0 \\
& & CnC(Proposed) & \textbf{99.8} & \textbf{99.9} & \textbf{1.0} \\
 \cline{2-6}
& \multirow{6}{*}{$\mathbf{LSUNr}$} & MSP  & 66.6 & 95.4 & 9.7  \\
& & ODIN \cite{liang2018enhancing} & 96.7 & 99.2 & 4.0  \\
& & Gen-ODIN \cite{hsu2020generalized} & 97.4 & 99.4 & 3.5  \\
& & Mahalanobis \cite{lee2018simple} & 88.0 & 95.8 & 8.4  \\
& & Gram Matrices \cite{sastry2020detecting} & 99.5  & 99.9 & 1.4 \\
& & CnC(Proposed) & \textbf{100.0} & \textbf{100.0} & \textbf{0.6} \\
 \cline{2-6}
& \multirow{6}{*}{$\mathbf{iSUN}$} & MSP  & 62.5 & 94.7 & 10.8  \\
& & ODIN \cite{liang2018enhancing} & 95.5 & 99.0 & 4.7  \\
& & Gen-ODIN \cite{hsu2020generalized}& 97.6 & 99.4 & 3.6  \\
& & Mahalanobis \cite{lee2018simple} & 87.0 & 96.0 & 8.8  \\
& & Gram Matrices \cite{sastry2020detecting} & 99.0  & 99.8 & 2.1 \\
& & CnC(Proposed) & \textbf{99.9} & \textbf{100.0} & \textbf{0.7} \\
  \cline{2-6}
& \multirow{6}{*}{$\mathbf{SVHN}$} & MSP  & 40.2 & 89.9 & 16.8  \\
& & ODIN \cite{liang2018enhancing} & 76.8 & 95.3 & 11.9  \\
& & Gen-ODIN \cite{hsu2020generalized} & 89.0 & 98.0 & 7.8  \\
& & Mahalanobis \cite{lee2018simple} & 83.2 & 93.0 & 10.7  \\
& & Gram Matrices \cite{sastry2020detecting} & 96.1  & 99.1 & 4.1 \\
& & CnC(Proposed) & \textbf{99.7} & \textbf{99.9} & \textbf{0.9} \\
\bottomrule
\end{tabular}
}
 \caption{Comparison with competitive \ood detection methods . We use DenseNet as network architecture for all methods. }
\label{tab:densenet10}
\end{table} 

\begin{table}
\centering

 \begin{tabular}{l|ll|ccc}
\toprule
 $\mathcal{D}_{in}^{train}$ & $\mathcal{D}_{out}^{test}$ & $\mathbf{Method}$ & $\mathbf{TNR}$ & $\mathbf{AUROC}$ & $\mathbf{DetErr}$\\ 
 
  & & & $\mathbf{(95\% TPR)}$ & & \\
  & & & $\pmb{\uparrow}$ & $\pmb{\uparrow}$ & $\pmb{\downarrow}$  \\ 
 \cline{2-6}
 
  \multirow{36}{*}{\begin{turn}{90}$\mathbf{CIFAR-100}$\end{turn}} & \multirow{6}{*}{$\mathbf{TINc}$} & MSP \cite{hendrycks2016baseline}  & 24.6 & 76.2 & 31  \\
& & ODIN \cite{liang2018enhancing} & 84.6 & 97.2 & 8.9  \\
& & Gen-ODIN \cite{hsu2020generalized} & 86.5 & 97.3 & 8.2 \\
& & Mahalanobis \cite{lee2018simple}& 53.8 & 93.1 & 15.1  \\
& & Gram Matrices  \cite{sastry2020detecting} & 89.0  & 97.7 & 7.5 \\
& & CnC(Proposed) & \textbf{99.7} & \textbf{99.9} & \textbf{1.3}\\
 \cline{2-6}
 & \multirow{6}{*}{$\mathbf{TINr}$} & MSP \cite{hendrycks2016baseline}  & 17.6 & 71.6 & 34.3  \\
& & ODIN \cite{liang2018enhancing} & 64.6 & 92.8 & 15.2  \\
& & Gen-ODIN \cite{hsu2020generalized}& 91.6 & 98.3 & 6.4  \\
& & Mahalanobis \cite{lee2018simple} & 79.1 & 93.6 & 11.7  \\
& & Gram Matrices  \cite{sastry2020detecting} & 95.7  & 99.0 & 4.5 \\
& & CnC(Proposed) & \textbf{99.6} & \textbf{99.8} & \textbf{1.6} \\
 \cline{2-6}
& \multirow{6}{*}{$\mathbf{LSUNc}$} & MSP \cite{hendrycks2016baseline}  & 28.6 & 80.2 & 27.3  \\
& & ODIN \cite{liang2018enhancing} & 84.6 & 97.3 & 8.7  \\
& & Gen-ODIN \cite{hsu2020generalized} & 77.5 & 95.8 & 11.1  \\
& & Mahalanobis \cite{lee2018simple} & 69.5 & 93.1 & 12.4  \\
& & Gram Matrices  \cite{sastry2020detecting} & 65.5  & 91.4 & 16.4 \\
& & CnC(Proposed) & \textbf{98.3} & \textbf{99.6} & \textbf{3.1} \\
 \cline{2-6}
& \multirow{6}{*}{$\mathbf{LSUNr}$} & MSP \cite{hendrycks2016baseline}  & 17.6 & 70.8 & 35  \\
& & ODIN \cite{liang2018enhancing}  & 65.2 & 93.4 & 14.2  \\
& & Gen-ODIN \cite{hsu2020generalized} & 91.4 & 98.2 & 6.3  \\
& & Mahalanobis \cite{lee2018simple} & 81 & 95.6 & 10.6 \\
& & Gram Matrices  \cite{sastry2020detecting} & 97.2  & 99.3 & 3.6 \\
& & CnC(Proposed) & \textbf{99.9} & \textbf{99.9} & \textbf{0.9} \\
 \cline{2-6}
& \multirow{6}{*}{$\mathbf{iSUN}$} & MSP \cite{hendrycks2016baseline}  & 14.9 & 69.5 & 36.2  \\
& & ODIN \cite{liang2018enhancing} & 61.1 & 92.1 & 19.6  \\
& & Gen-ODIN \cite{hsu2020generalized} & 91.2 & 98.3 & 6.6  \\
& & Mahalanobis \cite{lee2018simple} & 79.8 & 94.6 & 11.0  \\
& & Gram Matrices  \cite{sastry2020detecting} & 95.9  & 99.0 & 4.4 \\
& & CnC(Proposed) & \textbf{99.6} & \textbf{99.9} & \textbf{1.4} \\
 \cline{2-6}
& \multirow{6}{*}{$\mathbf{SVHN}$} & MSP \cite{hendrycks2016baseline} & 20.9 & 82.7 & 24.4  \\
& & ODIN \cite{liang2018enhancing} & 10.0 & 70.5 & 33.6  \\
& & Gen-ODIN \cite{hsu2020generalized} & 80.3 & 96.6 & 9.7  \\
& & Mahalanobis \cite{lee2018simple}& 46.5 & 86.3 & 19.6  \\
& & Gram Matrices  \cite{sastry2020detecting} & 89.3  & 97.3 & 7.6 \\
& & CnC(Proposed) & \textbf{98.7} & \textbf{99.7} & \textbf{2.0} \\
\bottomrule

 
\end{tabular}
 \caption{Comparison with competitive \ood detection methods . We use DenseNet as network architecture for all methods. }
\label{tab:densenet100}
\end{table}

\noindent\setlength\tabcolsep{6pt}%
\begin{table}[t]
	 \centering
\resizebox{0.9\linewidth}{!}{%
 \begin{tabular}{l|ll|ccc}
\toprule
 $\mathcal{D}_{in}^{train}$ & $\mathcal{D}_{out}^{test}$ & $\mathbf{Method}$ & $\mathbf{TNR}$ & $\mathbf{AUROC}$ & $\mathbf{DetErr}$\\ 
 
 &  & & $\mathbf{(95\% TPR)}$ & & \\
 &  & & $\pmb{\uparrow}$ & $\pmb{\uparrow}$ & $\pmb{\downarrow}$  \\ 
 \cline{2-6}
 
 \multirow{36}{*}{\begin{turn}{90}$\mathbf{CIFAR-10}$\end{turn}} & \multirow{6}{*}{$\mathbf{TINc}$} & MSP \cite{hendrycks2016baseline}  & 43.9 & 91.0 & 15.2  \\
& & ODIN \cite{liang2018enhancing} & 66.2 & 92.6 & 15.6  \\
& & Gen-ODIN \cite{hsu2020generalized} & 84.6 & 97.1 & 8.8  \\
& & Mahalanobis \cite{lee2018simple} & 76.0 & 95.0 & 12.6  \\
& & Gram Matrices \cite{sastry2020detecting} & 96.7  & 99.2 & 3.9 \\
& & CnC(Proposed) & \textbf{100.0} & \textbf{100.0} & \textbf{0.5}\\
  \cline{2-6}
& \multirow{6}{*}{$\mathbf{TINr}$} & MSP \cite{hendrycks2016baseline}  & 42.1 & 90.3 & 15.8  \\
& & ODIN \cite{liang2018enhancing}& 69.0 & 93.1 & 14.8  \\
& & Gen-ODIN \cite{hsu2020generalized} & 83.8 & 96.8 & 9.1  \\
& & Mahalanobis \cite{lee2018simple} & 85.5 & 96.8 & 9.3 \\
& & Gram Matrices \cite{sastry2020detecting} & 98.7  & 99.7 & 3.2 \\
& & CnC(Proposed) & \textbf{99.9} & \textbf{100.0} & \textbf{0.9} \\
  \cline{2-6}
& \multirow{6}{*}{$\mathbf{LSUNc}$} & MSP \cite{hendrycks2016baseline}  & 45.8 & 91.7 & 13.6  \\
& & ODIN \cite{liang2018enhancing}& 63.6 & 91.4 & 17.0  \\
& & Gen-ODIN \cite{hsu2020generalized} & 89.2 & 98.0 & 7.4 \\
& & Mahalanobis \cite{lee2018simple} & 70.4 & 93.0 & 14.7  \\
& & Gram Matrices \cite{sastry2020detecting} & 89.8  & 97.8 & 8.4 \\
& & CnC(Proposed) & \textbf{99.9} & \textbf{100.0} & \textbf{0.8} \\
  \cline{2-6}
& \multirow{6}{*}{$\mathbf{LSUNr}$} & MSP \cite{hendrycks2016baseline} & 41.2 & 90.1 & 15.8  \\
& & ODIN \cite{liang2018enhancing} & 70.6 & 93.2 & 14.4  \\
& & Gen-ODIN \cite{hsu2020generalized} & 92.9 & 98.5 & 5.8  \\
& & Mahalanobis \cite{lee2018simple} & 87.3 & 97.2 & 8.4  \\
& & Gram Matrices \cite{sastry2020detecting} & 98.7  & 99.7 & 1.4 \\
& & CnC(Proposed) & \textbf{100.0} & \textbf{100.0} & \textbf{0.7} \\
 \cline{2-6}
& \multirow{6}{*}{$\mathbf{iSUN}$} & MSP \cite{hendrycks2016baseline}  & 41.9 & 90.3 & 15.7  \\
& & ODIN \cite{liang2018enhancing}& 71.0 & 93.5 & 14.3  \\
& & Gen-ODIN \cite{hsu2020generalized} & 91.6 & 98.3 & 6.3  \\
& & Mahalanobis \cite{lee2018simple} & 86.0 & 97.0 & 9.1  \\
& & Gram Matrices \cite{sastry2020detecting} & 99.3  & 99.8 & 1.9 \\
& & CnC(Proposed) & \textbf{100.0} & \textbf{100.0} & \textbf{0.7} \\
 \cline{2-6}
& \multirow{6}{*}{$\mathbf{SVHN}$} & MSP \cite{hendrycks2016baseline}  & 27.9 & 89.3 & 14.9  \\
 & & ODIN \cite{liang2018enhancing}& 39.1 & 84.9 & 24.4  \\
 & & Gen-ODIN \cite{hsu2020generalized}& 90.8 & 98.4 & 6.6  \\
 & & Mahalanobis \cite{lee2018simple}  & 96.8 & 99.1 & 3.8  \\
 & & Gram Matrices \cite{sastry2020detecting} & 97.6  & 99.5 & 3.3 \\
 & & CnC(Proposed) & \textbf{99.6} & \textbf{99.9} & \textbf{1.2} \\
\bottomrule
\end{tabular}
}
 \caption{Comparison with competitive \ood detection methods . We use ResNet as network architecture for all methods. }
\label{tab:resnet10}
\end{table} 

\begin{table}[t]
	\centering
 
 \begin{tabular}{l|ll|ccc}
\toprule
 $\mathcal{D}_{in}^{train}$ & $\mathcal{D}_{out}^{test}$ & $\mathbf{Method}$ & $\mathbf{TNR}$ & $\mathbf{AUROC}$ & $\mathbf{DetErr}$\\ 
 
  & & & $\mathbf{(95\% TPR)}$ & & \\
  & & & $\pmb{\uparrow}$ & $\pmb{\uparrow}$ & $\pmb{\downarrow}$  \\ 
 \cline{2-6}

\multirow{36}{*}{\begin{turn}{90}$\mathbf{CIFAR-100}$\end{turn}} & \multirow{6}{*}{$\mathbf{TINc}$} & MSP \cite{hendrycks2016baseline} & 21.8 & 77.1 & 29.6  \\
& & ODIN \cite{liang2018enhancing}& 41.8 & 84.3 & 22.7  \\
& & Gen-ODIN \cite{hsu2020generalized}& 80.4 & 96.6 & 9.7   \\
& & Mahalanobis \cite{lee2018simple} & 49.8 & 85.0 & 23.0  \\
& & Gram Matrices \cite{sastry2020detecting} & 88.5  & 97.7 & 7.8 \\
& & CnC(Proposed) & \textbf{99.5} & \textbf{99.8} & \textbf{1.8}\\
 \cline{2-6}
& \multirow{6}{*}{$\mathbf{TINr}$} & MSP \cite{hendrycks2016baseline}  & 17.6 & 73.4 & 32.0  \\
& & ODIN \cite{liang2018enhancing} & 45.5 & 86.1 & 21.6  \\
& & Gen-ODIN \cite{hsu2020generalized}& 87.3 & 97.6 & 7.9  \\
& & Mahalanobis \cite{lee2018simple} & 59.0 & 88.1 & 19.6  \\
& & Gram Matrices \cite{sastry2020detecting} & 94.8  & 98.9 & 5.0 \\
& & CnC(Proposed) & \textbf{99.0} & \textbf{99.7} & \textbf{2.2} \\
 \cline{2-6}
& \multirow{6}{*}{$\mathbf{LSUNc}$} & MSP \cite{hendrycks2016baseline}  & 16.1 & 74.1 & 31.8  \\
& & ODIN \cite{liang2018enhancing} & 26.7 & 77.5 & 28.9  \\
& & Gen-ODIN \cite{hsu2020generalized}& 69.5 & 94.6 & 13  \\
& & Mahalanobis \cite{lee2018simple} & 37.0 & 80.2 & 28.4  \\
& & Gram Matrices \cite{sastry2020detecting} & 64.8  & 92.1 & 15.8 \\
& & CnC(Proposed) & \textbf{98.4} & \textbf{99.6} & \textbf{2.9} \\
 \cline{2-6}
& \multirow{6}{*}{$\mathbf{LSUNr}$} & MSP \cite{hendrycks2016baseline}  & 14.9 & 70.9 & 33.9 \\
& & ODIN \cite{liang2018enhancing} & 39.8 & 82.8 & 24.5  \\
& & Gen-ODIN \cite{hsu2020generalized}& 86.1 & 97.5 & 7.8  \\
& & Mahalanobis \cite{lee2018simple} & 56.5 & 91.0 & 17.3 \\
& & Gram Matrices \cite{sastry2020detecting} & 96.6  & 99.2 & 3.3 \\
& & CnC(Proposed) & \textbf{99.6} & \textbf{99.9} & \textbf{1.6} \\
 \cline{2-6}
& \multirow{6}{*}{$\mathbf{iSUN}$} & MSP \cite{hendrycks2016baseline}  & 15.1 & 72.6 & 32.7  \\
& & ODIN \cite{liang2018enhancing} & 43.8 & 84.5 & 22.9  \\
& & Gen-ODIN \cite{hsu2020generalized} & 84.6 & 97.1 & 8.8  \\
& & Mahalanobis \cite{lee2018simple} & 60.0 & 90.3 & 17.4  \\
& & Gram Matrices \cite{sastry2020detecting} & 94.8 & 98.8 & 4.4 \\
& & CnC(Proposed) & \textbf{98.6} & \textbf{99.6} & \textbf{2.4} \\
 \cline{2-6}
& \multirow{6}{*}{$\mathbf{SVHN}$} & MSP \cite{hendrycks2016baseline}  & 14.9 & 76.4 & 29.1  \\
& & ODIN \cite{liang2018enhancing} & 21.6 & 73.9 & 32.1  \\
& & Gen-ODIN \cite{hsu2020generalized}& 48.0 & 91.8 & 14.9  \\
& & Mahalanobis \cite{lee2018simple} & 33.8 & 86.0 & 21.3  \\
& & Gram Matrices \cite{sastry2020detecting} & 80.8  & 96.0 & 10.4 \\
& & CnC(Proposed) & \textbf{98.3} & \textbf{99.6} & \textbf{2.3} \\
\bottomrule
\end{tabular}

\caption{Comparison with competitive \ood detection methods . We use ResNet as network architecture for all methods. } 
\label{tab:resnet100}
\end{table}

\begin{table}
\centering
\centering
 
 \begin{tabular}{lccccc}
\toprule

  $\mathbf{Method}$ & $\mathbf{TNR}$ & $\mathbf{AUROC}$ & $\mathbf{DetErr}$ & $\mathbf{Mean}$ & $\mathbf{Mean}$\\ 
 
   &  $\mathbf{(95\% TPR)}$ & & &Diversity & Entropy \\
   &  $\pmb{\uparrow}$ & $\pmb{\uparrow}$ & $\pmb{\downarrow}$ & $\pmb{\uparrow}$ & $\pmb{\uparrow}$ \\ 
\toprule
  \textbf{CnC(proposed)}  & \textbf{98.3} & \textbf{99.6} & \textbf{2.6} & \textbf{3.4} & 0.4 \\
  \textbf{PBCC} & 93.7 & 98.6 & 6.2 & 2.27 & 0.33 \\
  \textbf{Corruptions} & 97.5 & 99.4 & 3.5 & 3.38 & 0.38 \\
  \textbf{AnC} & 70.2 & 94.4 & 14.1 & 1.57 & \textbf{0.92} \\
  \textbf{CnM} & 68.3 & 92.5 & 14.0 & 3.3 & 0.32 \\
\bottomrule

\end{tabular}
\caption{We aim to show that although adversarially perturbed images can cause a model to generate low confidence probability vectors for OOD class ( This means high entropy on samples that correspond to OOD) but generate sub-optimal results. Thus the higher entropy argument does not hold for adversarially perturbed images although the higher diversity leading to higher OOD detection performance holds true. \textbf{Experimental Setup :} We consider CIFAR-10 dataset as ID and a K-class classifier pretrained on CIFAR-10 as input. To ensure a controlled setting, we first generate and save PBCC images locally. Now we adversarially perturb PBCC images using PGD-L2 attack~\cite{madry2019deep} to ensure that they cause the model to generate low confidence probability vectors close to uniform distribution. We name this method of generating proxy \ood samples as \textbf{AnC (Adversarial Perturbations over PBCC)}
Comparison of OOD detection performance metrics CnC (Proposed), PBCC, Corruptions~\cite{hendrycks2018benchmarking}, AnC and CnM \ood detectors. Note CnM refers to Corruption over Mixup\cite{Pang*2020Mixup} augmentation technique (See description of CnM on Table \ref{tab:CnCvsCnM}). We use ResNet as network architecture, CIFAR-10 as ID dataset and TinyImageNet(Crop) as \ood dataset for all methods.}
\label{tab:densenetAnCCnM}
\end{table}

\section{More Comparison with State-of-the-art}
\label{sec:compWithSOA}
To demonstrate the effectiveness of CnC based data augmentation, we provide results of TNR@0.95TPR, AUROC and Detection Error of our method against the competing methods. We demonstrate that our method yields superior results on standard benchmark ID-OOD dataset pairs. To show the results are superior and also independent of feature extractor, we perform comparison using two standard feature extractors, DenseNet-BC and Resnet-34. Please refer to Table~\ref{tab:densenet10}, Table~\ref{tab:densenet100} for results obtained using DenseNet as feature extractor. In Table~\ref{tab:resnet10}, Table~\ref{tab:resnet100} we show results when ResNet-34 backbone is used for experimentation.

\begin{algorithm}[t]
	\label{algo_Rasha}
	\SetKwInput{KwInput}{Inputs}                
	\SetKwInput{KwOutput}{Output}  
	\DontPrintSemicolon
	\LinesNotNumbered
	\KwInput{
		$\mathcal{D}^{train}_{OOD}$  \tcp*{proxy OOD samples} 
		$F$ \tcp*{K class classifier pre-trained on $\mathcal{D}^{train}_{in}$ }
		 
		$Diversity$ \tcp*{training set diversity}
	}

	\KwOutput{Diversity $DataDiv$}
	 \SetKwFunction{FSum}{OodDiversity}
	
	\SetKwProg{Fn}{Function}{}{}
	
	\Fn{\FSum{$X$}}{
		\tcp{$X$ is an array of logits}
		$div \leftarrow$ $\emptyset$ \;
		\ForEach{$x_{i}$ in $X$}{
			l = $\min$ $d(x_{i},x_{j})$ \tcp{d(.) is distance metric, and $i\neq j$}
			$div \leftarrow$ \{$div \cup l$\}\;
		}\;

		\KwRet $average(div)$ 
	}
	 
	 \tcp{Main loop}
	 $X \leftarrow$ $\emptyset$ \;
     \ForEach{ $image$  in $\mathcal{D}^{train}_{\ood}$}{          
	        $out$ = F($image$) \;  	\tcp{out is the logits obtained having K dimensions}
	        $X \leftarrow$ \{$X \cup out$\}\;
		}\;
	  $DataDiv \leftarrow $\FSum{X}
	\caption{Measuring OOD diversity}
	\label{alg:OODDiv}
\end{algorithm}

\begin{figure*}[t]
	\begin{center}
		\includegraphics[width= 0.9\linewidth]{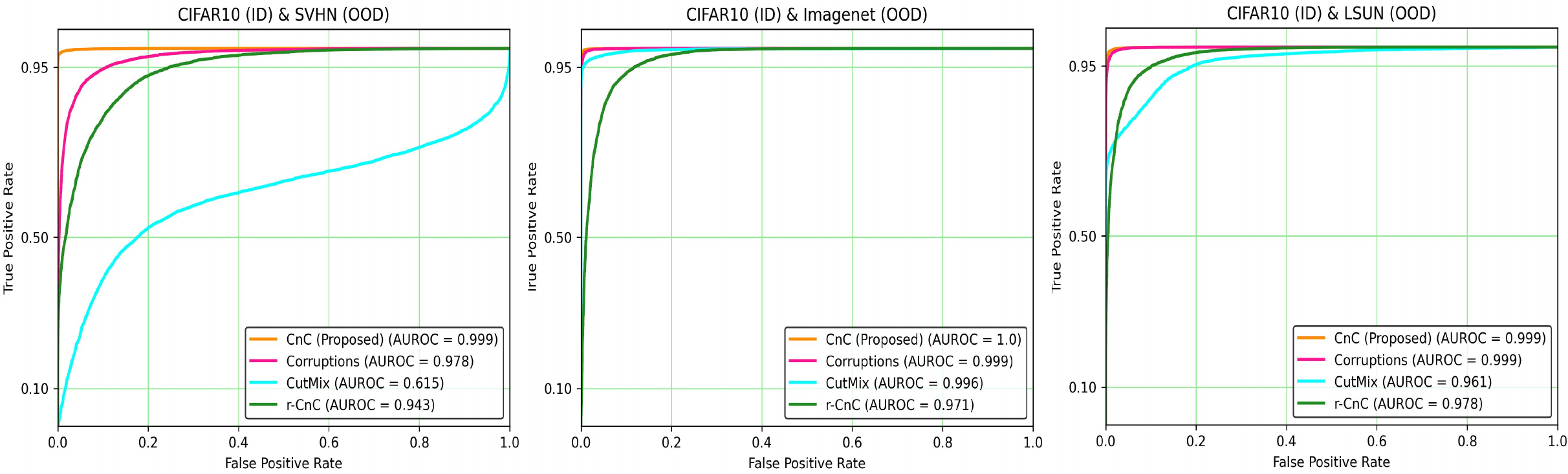}
		\caption{Comparison of ROC curves of CnC with PBCC, Corruptions~\cite{hendrycks2018benchmarking} and r-CnC on DenseNet-BC-100 network, where
			CIFAR-10 and SVHN, Imagenet, LSUN are in- and out-of-distribution datasets, respectively}
		\label{fig:auroc}
	\end{center}
\end{figure*}

\begin{table}[t]
 \centering

 \begin{tabular}{l|l|l|ccc}
\toprule
 
 $\mathcal{D}_{in}^{train}$ & $\mathcal{D}_{out}^{test}$ & $\mathbf{Method}$ & $\mathbf{TNR}$ & $\mathbf{AUROC}$ & $\mathbf{DetErr}$\\ 
 
  & & & $\mathbf{(95\% TPR)}$ & & \\
  & & & $\pmb{\uparrow}$ & $\pmb{\uparrow}$ & $\pmb{\downarrow}$  \\ 
 \cline{2-6}
 
 \multirow{12}{*}{\begin{turn}{90}$\mathbf{CIFAR-10}$\end{turn}} & \multirow{2}{*}{$\mathbf{TINc}$} & CnM  & 68.3 & 92.5 & 14.0  \\
& & CnC  & \textbf{100.0} & \textbf{100.0} & \textbf{0.5}  \\
 \cline{2-6} & \multirow{2}{*}{$\mathbf{TINr}$} & CnM  & 60.2 & 90.4 & 16.4  \\
& & CnC  & \textbf{99.9} & \textbf{100.0} & \textbf{0.9}  \\
 \cline{2-6}
& \multirow{2}{*}{$\mathbf{LSUNc}$} & CnM  & 86.7 & 96.8 & 8.6  \\
& & CnC  & \textbf{99.9} & \textbf{100.0} & \textbf{0.8}  \\
 \cline{2-6}
 & \multirow{2}{*}{$\mathbf{LSUNr}$} & CnM  & 71.1 & 93.5 & 13.0  \\
& & CnC  & \textbf{100.0} & \textbf{100.0} & \textbf{0.7}  \\
 \cline{2-6}
 & \multirow{2}{*}{$\mathbf{iSUN}$} & CnM  & 82.4 & 95.0 & 10.6  \\
& & CnC  & \textbf{100.0} & \textbf{100.0} & \textbf{0.7}  \\
 \cline{2-6}
 & \multirow{2}{*}{$\mathbf{SVHN}$} & CnM  & 56.7 & 93.8 & 11.9  \\
& & CnC  & \textbf{99.6} & \textbf{99.9} & \textbf{1.2}  \\
\bottomrule
\end{tabular}

 \caption{We show that Compounded Corruptions (Corruption Over PBCC (CnC)) generates better proxy \ood samples as compared to Corruption over Mixup~\cite{zhang2017mixup} (CnM). Mixup~\cite{zhang2017mixup} generates new data through convex combination of training samples and labels to improve DNN generalization. We hypothesize that PBCC generates more diverse samples and thus spans a larger region of the \ood space as compared to Mixup~\cite{zhang2017mixup}, hence, better exploration of the \ood space and better accuracy at test time. Using Algorithm~\ref{alg:OODDiv}, we find the mean diversity of CnM generated samples are $3.29$ where as that of CnC is $3.40$ (as indicated in Table~\ref{tab:densenetAnCCnM}). We use ResNet as network architecture for all methods.}
\label{tab:CnCvsCnM}
\end{table}

\clearpage

{\small

\bibliographystyle{splncs04}
\bibliography{main}
}

\end{document}